\documentclass{IEEEtran}

\usepackage{pgfplots}
\pgfplotsset{compat=1.16}
\pgfplotsset{every x tick label/.append style={font=\scriptsize, yshift=0.05cm}}
\pgfplotsset{every y tick label/.append style={font=\scriptsize, xshift=0.05cm}}

\usepackage{generic}
\usepackage{comment}
\usepackage{amsmath,amssymb,amsfonts}
\usepackage{algorithmic}
\usepackage{graphicx}
\usepackage{textcomp}
\usepackage{multicol,multienum, multirow}
\usepackage{threeparttable}
\usepackage[font=normalsize]{caption}
\usepackage{booktabs}
\usepackage{colortbl}
\usepackage[ruled]{algorithm2e}
\usepackage[super]{nth}

\usepackage{hyperref}
\hypersetup{
    colorlinks,
    citecolor=black,
    filecolor=black,
    linkcolor=black,
    urlcolor=black
}

\begin{document}

\title{ANOVA-based Automatic Attribute Selection and a Predictive Model for Heart Disease Prognosis}

\author{Mohammed Nowshad Ruhani Chowdhury,
Wandong Zhang, \IEEEmembership{Member, IEEE},\\ Thangarajah Akilan, \IEEEmembership{Member, IEEE}

\thanks{M.N-R. Chowdhury is with the Department of Computer Science and Engineering at Metropolitan University, Sylhet, Bangladesh (e-mail: nowshad.cse@metrouni.edu.bd). W. Zhang is with the Department of Electrical and Computer Engineering at University of Windsor, Windsor, Ontario, Canada (e-mail: zhang1lq@uwindsor.ca). T. Akilan is with the Department of Software Engineering at Lakehead University, Thunder Bay, Ontario, Canada  (e-mail: takilan@lakeheadu.ca).}}

\maketitle

\begin{abstract}

Studies show that cardiovascular diseases (CVDs) are malignant and these diseases can be the cause of premature death of human beings. Thus, it is crucial to have an efficient prognosis for CVDs. However, the prognosis of heart disease is a very challenging task in clinical practices. In response to this, the healthcare industry has adopted machine learning-based algorithms built using sample data of CVDs to alleviate the process. Nevertheless, existing algorithms do not pay much attention to the selection of appropriate attributes that can significantly improve the prediction accuracy of the models. Thus, this work proposes an information fusion technique that combines key attributes of a person through analysis of variance (ANOVA) and domain experts’ knowledge. It also introduces a new collection of CVD data samples for emerging research. There are thirty-eight experiments conducted exhaustively to verify the performance of the proposed framework on four publicly available benchmark datasets and the newly created dataset in this work. The ablation study shows that the proposed approach can achieve a competitive mean average accuracy (mAA) of $99.2\%$ and a mean average AUC of $97.9\%$.

\end{abstract}

\begin{IEEEkeywords}
Cardiovascular disease, statistical analysis, machine learning, feature selection.  
\end{IEEEkeywords}

\section{Introduction}\label{sec:introduction}

\begin{figure}[!htb]
\centerline{\includegraphics[trim={2.62cm, 2.5cm, 4.3cm, 2.5cm}, clip, width=1.0\columnwidth]{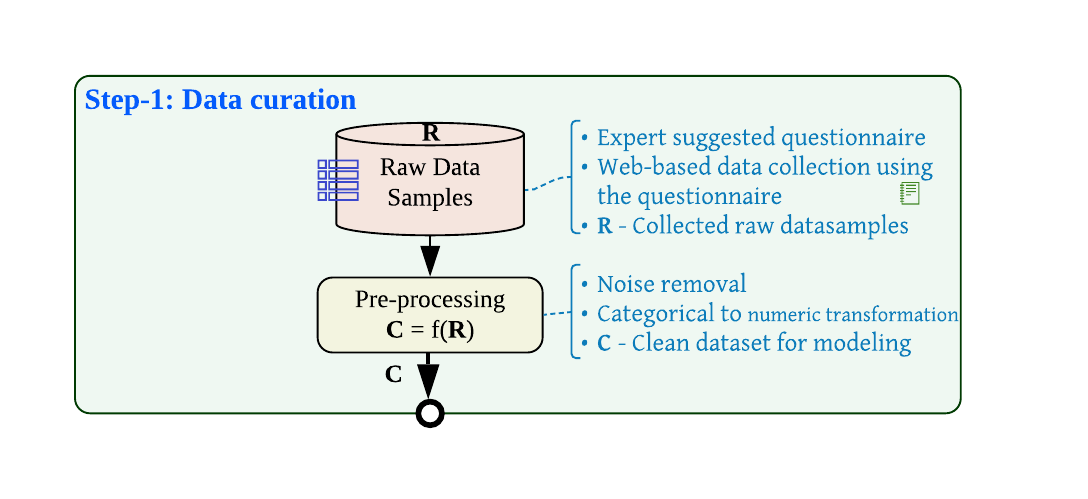}}
\centerline{\includegraphics[trim={2.62cm, 2.5cm, 2.62cm, 2.5cm}, clip, width=1.0\columnwidth]{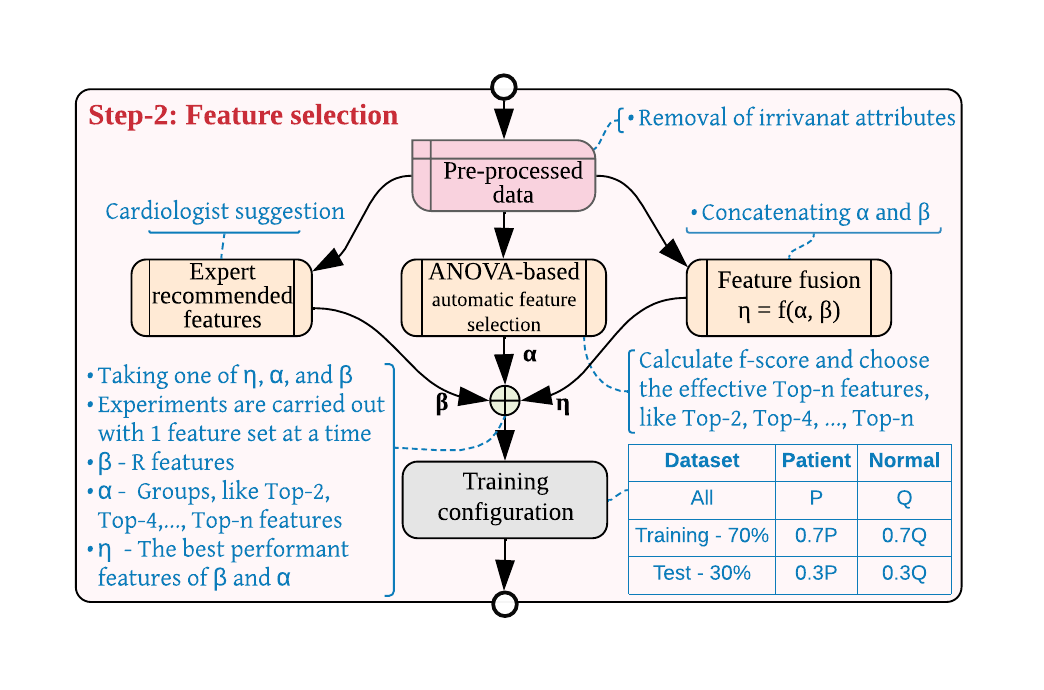}}
\centerline{\includegraphics[trim={2.55cm, 2.5cm, 2.62cm, 2.5cm}, clip, width=\columnwidth]{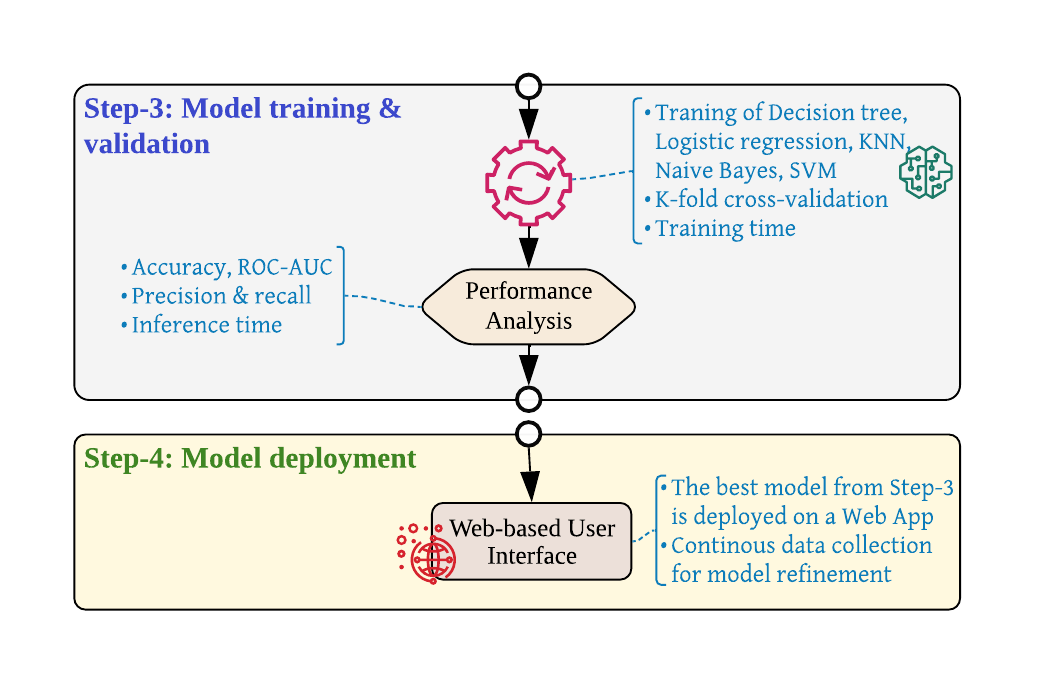}}
\caption{Overview of the Proposed Smart Heart Disease Prognosis Pipeline with Automatic Attribute Selection and Fusion.}
\vspace{-0.2cm}
\label{flowdiagram}
\end{figure}

\IEEEPARstart{C}{ardiovascular} diseases, a.k.a heart diseases, are one of the primary reasons for people's death, as WHO estimates 17.9 million people die every year due to this. In other words, it is the cause of 32 percent of worldwide mortality\footnote[1]{\url{https://www.who.int/health-topics/cardiovascular-diseases##tab=tab_1}}, and this rate is increasing dramatically\footnote[2]{\url{https://www.who.int/news-room/fact-sheets/detail/the-top-10-causes-of-death}}. Unfortunately, it is found that CVDs are also paramount causes of many premature deaths. Studies show that approximately one-third of deaths occur in people under the age of 70 due to CVDs\footnotemark[1]. It mainly affects the average life expectancy in the developing nations\footnote[3]{\url{https://www.who.int/news-room/fact-sheets/detail/cardiovascular-diseases-(cvds)}}, where about 75 percent of CVD-related deaths occur in the families having an annual income between low and average. 
\begin{table*}[tp]
\caption{Summary of the Literature Reviewed in Section~\ref{literature_review}}
    \label{tab:summary_literature}
    \setlength{\tabcolsep}{6pt}
    \centering
    \begin{tabular}{|m{1.7cm}|m{6.8cm}|m{1.9cm}|m{3.5cm}|m{2.2cm}|}
    \hline \hline
        \textbf{Dataset} $\rightarrow$ & \centering Cleveland & \centering Switzerland  & \centering Hungarian & \quad Long Beach Va\\
       \hline
       \textbf{The study that uses the dataset} $\rightarrow$ &  Mohan~\textit{et al.}~\cite{mohan2019effective},
       Shah~\textit{et al.}~\cite{shah2020heart},
       Uyar~\textit{et al.}~\cite {UYAR2017588},\newline
       El~Hamdaoui~\textit{et al.}~\cite{el2020clinical},
       Bashir~\textit{et al.}~\cite{bashir2014ensemble},
       Ganesan~\textit{et al.}~\cite{ganesan2019iot},
       Valarmathi~\textit{et al.}~\cite{valarmathi2021heart},
       Faiayaz~\textit{et al.}~\cite{FAIAYAZWARIS2021}, Ali~\textit{et al.}~\cite{ali2021heart}, \newline
       Amin~\textit{et al.}~\cite{amin2019identification},
       Gárate-Escamila~\textit{et al.}~\cite{garate2020classification},\newline
       El-Bialy~\textit{et al.}~\cite{el2015feature},
       Babu~\textit{et al.}~\cite{babu2018medical},
       Anooj~\textit{et al.}~\cite{anooj2012clinical}
       & 
       Babu~\textit{et al.}~\cite{babu2018medical},\newline
       Anooj~\textit{et al.}~\cite{anooj2012clinical}
       & 
       Gárate-Escamila~\textit{et al.}~\cite{garate2020classification},\newline
       El-Bialy~\textit{et al.}~\cite{el2015feature},\newline
       Babu~\textit{et al.}~\cite{babu2018medical}, Ali~\textit{et al.}~\cite{ali2021heart},
       Anooj~\textit{et al.}~\cite{anooj2012clinical}
       & El-Bialy~\textit{et al.}~\cite{el2015feature}
       \\
       \hline \hline
    \end{tabular}
\end{table*}

Therefore, cardiovascular disease prediction using large data analytical approach has become an essential part of medical practices for a better healthcare system. However, for building data-driven models, the availability of the required large data in the clinical domain is insufficient due to the following reasons: (i) the acceptance of patients to participate in the data collection process is very low, and (ii) the data collection process may include observation of participants' complex physical activities. Besides, to build an intelligent prognosis model for CVDs, it is important to select an appropriate set of patients' attributes. Where, the existing solutions use only the expert-suggested attributes; thus, their performances may not be optimized. To address these issues, this work: (i) introduces a new dataset, (ii) proposes a feature fusion technique that exploits expert-suggested attributes and ANOVA-based automatically distilled attributes, and (iii) implements an efficient online CVD prognosis pipeline.  

In this direction, we build five binary classifiers using well-known machine learning (ML) algorithms, viz. logistic regression (LR), decision tree (DT), k-nearest neighbor (KNN), naïve Bayes (NB), and support vector machine (SVM). These classifiers are tested on the proposed CVD prognosis pipeline shown in Fig.~\ref{flowdiagram}. The ablation study uses the newly created dataset named Bangladesh heart disease collection (BHDC) and four other publicly available benchmark datasets from the UCI machine learning repository that is available at \url{https://archive.ics.uci.edu/ml/datasets/heart+disease/}. Through exhaustive experiments (38 in total) using K-fold cross-validation method, the models' performances are evaluated using the five evaluation metrics. 

The rest of this article elaborates the conducted research under the following organization -- Section~\ref{literature_review} discusses the related works, Section~\ref{sec-proposedmodel} describes the proposed CVD prognosis pipeline step-by-step, Section~\ref{sec:experiments} provides a thorough analysis on the experimental findings, and finally, Section~\ref{sec:conclusion} concludes the article with future direction. 

\section{Literature Review}\label{literature_review}

Table~\ref{tab:summary_literature} summarizes the important research works carried out by the active ML community for heart disease prognosis, which are reviewed in this section. It must be mentioned that, unfortunately, most of the existing works do not perform a comparative analysis on all the datasets listed in Table~\ref{tab:summary_literature}. However, we have conducted a thorough experimental analysis using all the datasets, including the new dataset introduced in this work. Hence, we generalize the existing solutions into multi-modality approaches and uni-modality approaches, for organizational purpose.  

\subsection{Multi-modality Approaches}

Recently, researchers have focused on building hybrid models by integrating various linear and non-linear classifiers for producing improved predictions. For example, Mohan~\textit{et al.}~\cite{mohan2019effective} have proposed a hybrid model, named hybrid random forest with a linear model (HRFLM) that combines several classical ML algorithms, such as naïve Bayes, random forest, logistic regression, gradient boosted trees, support vector machine, and generalized linear models. Although this high complex integration is expected to deliver better results, they record a mere good accuracy of 88.7$\%$ on the pre-processed Cleveland dataset (cf.~row\#2,~Table~\ref{tab:datasets_summary}). 
Similarly, Uyar and İlhan~\cite {UYAR2017588} also propose a hybrid model that incorporates a genetic algorithm (GA) and recurrent fuzzy neural networks (RFNN) to diagnose heart diseases. Nonetheless, they report their best result of 97.8$\%$ accuracy on the same Cleveland dataset, they achieve this result on 15$\%$ of the entire samples, while they use 85$\%$ of the data for training. We think that validating the model's performance on just 15$\%$ of the samples is a very limited experimental study and it can not be conclusive. Likewise, Amin~\textit{et al.}~\cite{amin2019identification} present a dual model combination using naïve Bayes and logistic regression under a voting scheme. They verify this scheme on the Cleaveland dataset and achieve the best accuracy of 87.4$\%$.
Hence, Babu~\textit{et al.}~\cite{babu2018medical} built a neural network-based multi-model solution using grey wolf optimization (GWO) and an autoencoder-based recurrent neural network (RNN) for identifying heart disease patients. They employ the GWO for irrelevant and redundant attribute removal, and use an RNN for the classification purpose. They test this method on three UCI datasets, namely, Cleveland, Switzerland, and Hungarian (cf.~row\#2-4,~Table~\ref{tab:datasets_summary}). They show that the removal of unwanted features improves the classification performance by average of 10$\%$ compared to the baseline model without feature selection. 

\setlength\tabcolsep{6pt}
\begin{table*}[!t]
    \caption{Dataset Summary}
    \label{tab:datasets_summary}
    \centering
    \begin{tabular}{|p{3cm}|p{1.2cm}|p{1.2cm}|p{4.0cm}|p{6.0cm}|}
    \hline\hline
        \centering \multirow{2}{*}{\textbf{Dataset name}} & \centering \textbf{$\#$ of Samples} & \centering \textbf{$\#$ of Attributes$^\diamond$} & \centering \textbf{Expert Suggested Attributes' IDs for Predictive Modeling} & \multirow{2}{*}{\textbf{Comments}} \\
        \hline\hline
        BHDC & \centering 551 & \centering 18 & 1, 2, 3, 7, 8, 9, 10, 14 & For attribute details, refer to Table~\ref{tab-ourdatasets-attribute} in Section~\ref{sec-proposedmodel}\\\hline
        
        Cleveland$^*$ & \centering 282 & \centering 49 & 
        \multirow{4}{4cm}{3, 4, 9, 10, 12, 16, 19, 32, 38, 40, 41, 44, 51, 58}  & \multirow{4}{*}{The raw datasets has 76 attributes$^*$} \\\cline{1-3} 
        
        Hungarian$^*$ & \centering 294 & \centering 46 &  & \\\cline{1-3}  
         
        Switzerland$^*$ & \centering 123 & \centering 47 &  & \\\cline{1-3} 
        
        Long Beach VA$^*$ & \centering 200 & \centering 52 & & \\\cline{1-3}\hline\hline 
    \end{tabular}
    
    \noindent \footnotesize{$^*$detail of the attributes and their index information can be found at \url{https://archive.ics.uci.edu/ml/datasets/heart+disease/}\hfill \\
    \hspace{0.4cm} $^\diamond$number of attributes after removal of noisy or missing valued fields \hfill}
    \vspace{-0.2cm}
\end{table*}

\subsection{Uni-modality Approaches}
To handle CVD prognosis problem, some research works focus on building strong uni-modal classifiers, instead of applying multi-modality-based approaches. For instance, Valarmathi and Sheela~\cite{valarmathi2021heart} exploit a tree-based pipeline optimization tool (TPOT) based on GA for hyperparameter tuning of various classifiers. Using this technique, they optimize the hyper-parameters of random forest and XGBoost classifiers to achieve the best performance. They report the highest accuracy of 97.5$\%$ on the Cleveland dataset employing random forest classifier.
Similarly, Gárate-Escamila~\textit{et al.}~\cite{garate2020classification} study the impact of feature selection methods: chi-square, principal component analysis (PCA), and combination of chi-square and PCA (CHI-PCA) on a variety of classification algorithms. They report that a random forests (RF) classifier using CHI-PCA feature selection achieves the best results of 99\% average accuracy on Cleveland and Hungarian datasets. 
Likewise, El-Bialy~\textit{et al.}~\cite{el2015feature} have proposed a strategy using complementary information derived from different heart disease datasets. With this idea, they train a decision tree model on three UCI datasets, namely Cleveland, Hungarian, and Long Beach Va. Their approach achieves the best result of 78.5$\%$ accuracy. 

Hence, Anooj~\cite{anooj2012clinical} builds a custom clinical decision support model for the prediction of heart disease risk level using weighted fuzzy rules. The author, firstly, performs weighted feature selection to obtain the weighted fuzzy rules. Then, constructs a decision tree-based fuzzy system with respect to the weighted fuzzy rules and selected features. Since there is uncertainty in the fuzzy rules, the proposed model does not achieve excellent performance. It records the best result of 76.5$\%$ accuracy on the Cleveland dataset. Similarly, there are many other researchers, like  El Hamdaoui~\textit{et al.}~\cite{el2020clinical}, Bashir~\textit{et al.}~\cite{bashir2014ensemble}, Ganesan~\textit{et al.}~\cite{ganesan2019iot}, Shah~\textit{et al.}~\cite{shah2020heart},  Faiayaz~\textit{et al.}~\cite{FAIAYAZWARIS2021}, and Ali~\textit{et al.}~\cite{ali2021heart} also have focussed on building uni-modality-based  ML models, for CVD predictions. However, their approaches do not achieve significance performances.

Through above careful investigation, we find that the existing works do not systematically analyse the importance of key feature selection to improve the prediction time without compromising the classification accuracy. Besides, they have not tested their solutions on all UCI benchmark datasets, rather completed their studies on a limited number of datasets. Thus, this work rightly addresses these lackings in the existing works through a feature fusion technique exploiting expert-suggested attributes and ANOVA-based automatically selected attributes, and conducts an exhaustive experimental study on all UCI datasets and also on our own dataset (BHDC). The proposed pipeline mainly aims at reducing the cost and improving the prognosis of heart disease quickly and effectively.

\section{Proposed Model}\label{sec-proposedmodel}

The proposed automatic heart disease prognosis pipeline is illustrated in Fig.~\ref{flowdiagram}. It subsumes four key stages, namely, data curation (Section~\ref{data-curation}), feature selection (Section~\ref{feature-selection}), model training and validation (Section~\ref{model-training}), and Model deployment (Section~\ref{model-deployment}). The following subsections elaborates these stages step-by-step.
 
\subsection{Data Curation}\label{data-curation}

Data curation plays a vital role in ML-based analysis. It includes data collection, data cleaning, and data conditioning or transformation (cf.~Step-1, in Fig.~\ref{flowdiagram}). This study considers various sources for data gathering: its own dataset, coined as Bangladesh heart disease collection (BHDC), and the benchmark UCI ML repository\footnote[5]{\url{https://archive.ics.uci.edu/ml/datasets/heart+disease/}}. 

\subsubsection{Bangladesh Heart Disease Collection a.k.a. BHDC}\label{sec-BHDC}

Through a set of eighteen questions, focusing on personal and family attributes, this study gathered 564 data samples consisting of heart patients and normal people from six different hospitals and healthcare facilities located in Bangladesh. The questionnaire was prepared with consultation of two Cardiologists (i.e., domain experts), Dr. Muhammad Shahabuddin and Dr. Md. Anamur Rahman Anam\footnote[6]{\url{https://specialistdoctorsbd.com/cardiologist-sylhet-heart-specialist/}}. Dr. M. Shahabuddin is an Assistant Professor and the Head of the Department of Cardiology, Sylhet MAG Osmani Medical College. He is also a Clinical and Interventional Cardiologist at the Mount Adora Hospital, Akhalia, Sylhet, Bangladesh. Dr. Md. A. Rahman Anam is also a Cardiology Specialist at the Mount Adora Hospital, Akhalia, Sylhet, Bangladesh.

The six institutions involved in the data collection as follows: 1. Jalalabad Ragib-Rabeya Medical College and Hospital, Sylhet, 2. Mount Adora Hospital, Sylhet, 3. North East Medical College and Hospital, Sylhet, 4. Sylhet Women’s Medical College and Hospital, 5. National Heart Foundation Hospital, Sylhet, and 6. Sylhet MAG Osmani Medical College Hospital. After collecting the data samples, it is found that thirteen of them are outliers, i.e., they are not related to heart disease, like non-cardiac chest pain (NCCP). Such samples are removed, and the total number of legitimate samples is reduced to 551, representing 313 of heart disease patients and the rest of them are normal people (cf.~Table~\ref{tab:datasets_summary}). Finally, the raw categorical features are conditioned to have numerical representation and expert suggested labels are assigned to them, as listed in Table~\ref{tab-ourdatasets-attribute}. 

\subsubsection{UCI ML Repository}\label{sec-ucis'}

For extended experiments, we use four additional benchmark datasets from the UCI ML repository, namely Cleveland, Hungarian, Switzerland, and Long Beach Va. These datasets have conditioned attributes, i.e., attributes of the data samples are preprocessed to have numeric representation. However, some of the records contain noise. After removing such noisy records, there are 282, 294, 123, and 200 data samples in the Cleveland, Hungarian, Switzerland, and Long Beach datasets, respectively (cf.~Table~\ref{tab:datasets_summary}).

\begin{table}[!htb]
\caption{The Attribute Details of the BHDC Dataset}
\label{tab-ourdatasets-attribute}
\setlength{\tabcolsep}{3pt}
\begin{tabular}{|p{10pt} |p{65pt}|p{160pt}|}
\hline
$\#$&
\textbf{Attribute Name} & 
\textbf{Description of the Categorical Label}\\
\hline\hline
\multicolumn{3}{|c|}{Personal Attributes}\\ \hline
1 & Age & Age range - \{0: age $\leq$ 30, 1: 31$\leq$ age $\leq$ 45, 2: 46 $\leq$ age $\leq$ 65, 3 : age $\geq$ 65\} \\
2 & Gender & 0: Male, 1: Female, 2: Other\\
3 & Smoking habit & 0: No, 1: Yes, 2: N/A\\
4 & Smoking condition & 0: Past smoker, 1: Present smoker, 2: N/A\\
5 & Regular pulse & 0: No, 1: Yes , 2: N/A \\
6 & Physical activity & 0: Walk, 1: Outdoor games, 2: Gym , 3: No\\
7 & Diabetes & 0: No, 1: Yes, 2: N/A\\
8 & Cholesterol & 0: No, 1: Yes, 2: Unknown\\
9 & Chest pain & 0: Typical angina, 1: Atypical angina, 2: Non-cardiac chest
pain, 3: No chest pain \\
10 & Hypertension & 0: No, 1: Yes, 2: N/A\\
11 & Skipping doses & Habit of skipping heart failure medicines when
feeling better - \{0: No, 1: Yes, 2: N/A\}\\
12 & Junk food & Habit of eating junk food - \{0: Everyday, 1: 2-3 days a week, 2: 4-6 days a week, 3: No\}\\
13 & Rice eating habit & 0: 3 or more time a day, 1: 2 time a day, 2: 1 time a day\\ \hline
\multicolumn{3}{|c|}{Family Attributes}\\ \hline
1 4& Family history & Any family
member had/has heart disease - \{0: No, 1: Yes, 2: N/A\}\\
15 & Relationship & Relationship with the family member who had/has heart disease - \{0: Father and mother, 1: Brother and sister, 2: N/A\}\\
16 & Family member's age & Age of that family member who had/has heart disease - \{0: Less than 65, 1: Greater than 65, 2: N/A\}\\
17 & Family member's gender & Gender of that family member who had/has heart disease - \{0: Male, 1: Female, 2: N/A\}\\
18 & Disease type & Disease type of that family member - \{0: Heart Attack, 1: Heart Block, 2: N/A\}\\
19 & Label & Heart patient - \{0: No, 1: Yes\}\\
\hline\hline
\end{tabular}
\# - Attribute index
\vspace{-0.5cm}
\end{table}

\subsection{Feature Selection}\label{feature-selection}

Although each dataset consists of several attributes, for example, BHDC has 22 raw attributes, while all UCI datasets have 76 raw attributes. However, it is found that not all the attributes have significant correlation to the heart disease. Hence, it is important to identify the most informative set of attributes before modeling a predictive algorithm, to achieve a compute and memory efficient solution. For instance, the person's name, phone number, and address are not relevant features for classifying the person as heart patient or not. Thus, from each dataset, a set of core attributes are distilled. Here, for BHDC, Cleveland, Hungarian, Switzerland, and Long Beach Va, 18, 49, 46, 47, and 52 attributes are considered, respectively. 
This feature selection is further refined, through domain expertise, statistical analysis, and the combination of the both, as depicted in Fig.~\ref{flowdiagram}, Step-2, where, the  statistical-based selected features, expert recommended features, and the fusion of the both feature sets are denoted as $\alpha$, $\beta$, and $\eta$, respectively. For statistical-based automatic selection, ANOVA is employed, in this work.

\subsubsection{Expert Recommended Feature Selection}\label{Expert-Recommended-feature-selection}

This research collaborates with two experts from the clinical domain, Dr. Muhammad Shahabuddin and Dr. Md. Anamur Rahman Anam as stated in Section~\ref{sec-BHDC}.
These experts suggested to use eight attributes of the BHDC dataset. In Table~\ref{tab-ourdatasets-attribute}, these expert recommended features have the following attribute indices: \{1, 2, 3, 7, 8, 9, 10, 14\}. 
On the other hand, the UCI ML repository has listed its own set of fourteen recommended attributes. The details of the recommended attributes, for the UCI datasets, can be found in their attribute information table at \url{https://archive.ics.uci.edu/ml/datasets/heart+disease/}.

\subsubsection{ANOVA-based Automatic Feature Selection}\label{ANOVA-based-feature-selection}

The statistical approach, ANOVA is employed to identify the $n$-most impactful attributes from the raw datasets. It estimates the individual attribute's predictive strength based on their within class or intra-class and between class or inter-class discriminative ability, using mean and variance as expressed in \eqref{ee2}.
\begin{equation}
\begin{split}
&S^2_{between}=\frac{1}{C-1}\sum_{c=1}^{C}a_c(\bar X_c-\bar X)^2,\\
&S^2_{within}=\frac{1}{N-C}\sum_{c=1}^{C}\sum_{j=1}^{a_c}(X_{c,j}-\bar X_c)^2,
\label{ee2}
\end{split}
\end{equation}
where $S^2_{between}$, and $S^2_{within}$ refer to between groups sum of squared differences and within group sum of squared differences, respectively. Hence, $N$, $C$, $a_c$, $\bar X$, $\bar X_c$, and $X_{c,j}$ stand for the total number of samples across all the classes, the total number of classes, the number of samples belonging to the $c$-th class, the overall mean of the attribute across all classes, mean value of the attribute having class label $c$, and $j$-th attribute value belong to $c$-th class, respectively. 
In this study, the number of classes, $C$ is 2 (heart disease, normal), and the number of input attributes varies depending on the dataset under the experimental study. For every attribute, the algorithm computes $S^2_{between}$ and $S^2_{within}$, independently, followed by F-test statistic computation, as a ratio between these two quantities, using \eqref{e_2}.
\begin{equation}
\text{F-test} =\frac{S^2_{between}}{S^2_{within}}.
\label{e_2}
\end{equation}
The raw features are ranked based on computed F-test scores. In this way, we thoroughly analyse the impact of significant attribute sets by selecting Top-$n$, say Top-2, Top-4, and so forth feature sets. Note that the larger the F-test score an attribute gets, the higher the significance of it~\cite{ding2014identification, 9425544}.

\begin{figure}[!t]
\centerline{\includegraphics[trim={0.1cm, 0.1cm, 0.1cm, 0.1cm}, clip, width=0.5\columnwidth]{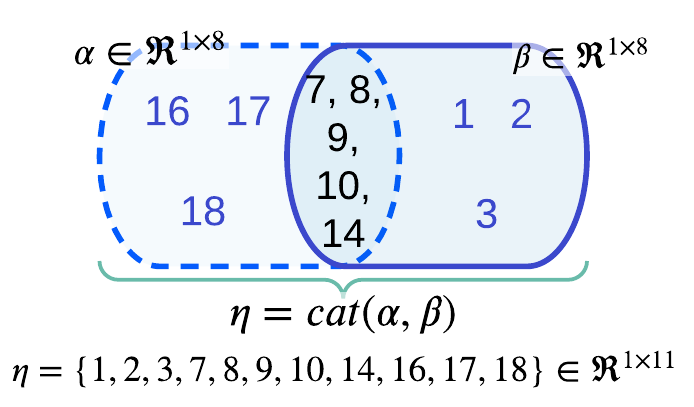}}
\caption{A Feature Fusion Example, Where Expert Recommended and ANOVA-Based Top-8 Features are Combined.}
\label{featurefusion}
\vspace{-0.2cm}
\end{figure}

\subsubsection{Feature Fusion}\label{Feature-Fusion-feature-selection}

As we have access to the expert recommended feature set, $\beta$ and ANOVA-based automatically selected feature set, $\alpha$ (cf.~Fig.~\ref{flowdiagram}), we generate a new group of features, $\eta$ by combining the above two sets, as expressed by \eqref{eq.fusion}.  
\begin{equation}
    \eta = cat(\alpha, \beta), 
    \label{eq.fusion}
\end{equation}
where $cat(\cdot)$ performs a non-overlapping feature concatenation of the feature sets $\alpha$ and $\beta$, such that the resultant, $\eta$ contains non-repeated unique set of features. 
For example, consider that BHDC dataset and its attribute indices tabulated in Table~\ref{tab-ourdatasets-attribute}. For BHDC, the experts suggested eight key attributes having the attribute indices: \{1, 2, 3, 7, 8, 9, 10, 14\}, while the ANOVA identifies the following as part of Top-8 features with the attribute indices: \{7, 8, 9, 10, 14, 16, 17, 18\}. Then, the fusion of these two sets using \eqref{eq.fusion} results an eleven-dimensional unique set of features belong to the attribute indices: \{1, 2, 3, 7, 8, 9, 10, 14, 16, 17, 18\}, as illustrated in Fig.~\ref{featurefusion}.

\subsection{Model Training and Validation }\label{model-training} 

Once the key attributes are identified, the samples of each dataset are split into mutually exclusive training and test sets, respectively taking 70$\%$ and 30$\%$ of the total number of samples. For example, in the BHDC dataset, there are 313 records of heart disease patients and 250 records of normal people. So, the training set gets 217 samples for heart disease conditions and 177 samples for normal conditions, while the test set gets 96, and 73 samples for heart disease and normal conditions, respectively. Based on these training and test sets, five different ML models: Decision tree, Logistic regression, KNN, naïve Bayes, and SVM are trained and their performances are validated (cf.~Fig.~\ref{flowdiagram}~-~Step~3). Extensive experiments are conducted to pick the best performant model with respect to various feature selection approaches discussed in Section~\ref{feature-selection}. The models' performances are measured using the important evaluation metrics, viz. accuracy, precision, recall ROC and AUC, and inference time.   

The following subsections elaborate the ML models used in this work.

\subsubsection{Logistic Regression} \label{lr}

The model selection begins with experiments on standard logistic regression ($\sigma(\cdot)$), where it is built upon an assumption that there is a non-linear (sigmoidal) relationship between the predictors and a binary target variable. 
It is known to be an effective classifier for dichotomous classification problems, since it produces output values in a bounded range of [0, 1] as defined by \eqref{eq.lr-Function}.

\begin{minipage}{0.45\columnwidth}
\begin{tikzpicture}[declare function={sigma(\x)=1/(1+exp(-\x));}]
\begin{axis}%
[
    width=2in,
    height=1.0in,
    grid=major,     
    xmin=-5,
    xmax=5,
    axis x line=bottom,
    ytick={0,.5,1},
    xtick={-5,5},
    xlabel={$z$},
    xlabel shift = -0.0cm,
    ymax=1,
    axis y line=middle,
    samples=10,
    domain=-5:5,
    legend style={at={(1,0.0)}, draw=none, fill=none, font =\scriptsize}   
]
    \addplot[blue,mark=none]   (x,{sigma(x)});
    \legend{$\sigma(z)$}
\end{axis}
\end{tikzpicture}

\end{minipage}%
\hfill
\begin{minipage}{0.45\columnwidth}

\begin{equation}
    \sigma(z) = \frac{1}{1+e^{-z}}, 
    \label{eq.lr-Function}
\end{equation}

\end{minipage}%

\noindent where $0 \leq \sigma(z) \leq 1$, and $z$ is an unbounded continuous value. The value, $z$ can be modeled as a linear combination of several weighted input attributes, like $z = X\cdot\omega$, where $\omega$ is a vector of learned parameters associated to the attribute vector, $X$. Thus, the score of the logistic regression can be interpreted as a probability of the positive outcome of an event to occur based on the observation, $X$, say presence of heart disease given a data sample with a set of patient's attributes. 
In this work, we used {sklearn}\footnote[7]{\href{https://scikit-learn.org/stable/modules/generated/sklearn.linear_model.LogisticRegression.html}{scikit-learn logistic regression}} LR library to create the logistic regression models, where the objective function was set to the default $\ell_2$ loss and the optimizer was selected through experiments. In this case, among \texttt{newton-cg}, \texttt{lbfgs}, and \texttt{liblinear} solvers, the \texttt{newton-cg} found to be best optimizer that produces the best classification performance with a fast convergence rate.

\subsubsection{Decision Tree}\label{dt}

In this work, the iterative dichotomiser 3 (ID3) algorithm was used for building DT classifiers. The ID3 computes Gini index as an attribute selection measure (ASM) for each level of the decision tree ($T$) recursively to build a strong classifier. Gini index defined in \eqref{eq-geni} is computationally efficient than its counterpart, information gain.  
\begin{equation}
    Geni(T) = 1 - \sum_{i=1}^{n} P^2(c_i),
    \label{eq-geni}
\end{equation}
where $P(c_i)$ is the probability of a randomly selected sample to have the classification label, $c_i$ in a $n$-tuple dataset. It is used to measure the impurity of the current samples, such that: $\forall_i \in \{1, 2, \cdots, n\}: p(c_i) \geq 0$ and $\sum_{i=1}^{n} P(c_i) = 1$. During the tree construction, an attribute with a lower Gini index is opted for a split.
In this work, the DT model is generate using
the \texttt{sklearn}\footnote[8]{\href{https://scikit-learn.org/stable/modules/generated/sklearn.tree.DecisionTreeClassifier.html}{scikit-learn decision tree classifier}} DT classifier library, where the \texttt{max-depth} was empirically selected between 3 and 6.

\subsubsection{K-Nearest Neighbour} \label{knn}

KNN is a non-parametric algorithm that classifies the samples based on the characteristic similarity measure between the test case and the training cases. 
In this study, the KNN classifiers are built using Euclidean distance as a similarity measure defined in~\eqref{eq.Euclidean}
\begin{equation}
    d(p,q) = \sqrt{\sum_{i=1}^{n}(q_i - p_i)^2}
    \label{eq.Euclidean}
\end{equation}
where, $p$ and $q$ are two data samples in an $n$-dimensional Euclidean space, $q_i$ and $p_i$ are Euclidean vectors, starting from the origin of the space, and $d(p,q)$ is the Euclidean distance between the two data samples represented in the Euclidean space. We implement the KNN-based CVDs classifiers using the off-the-shelf \texttt{sklearn} KNeighbors classifier library\footnote[9]{\href{https://scikit-learn.org/stable/modules/generated/sklearn.neighbors.KNeighborsClassifier.html}{scikit-learn K-Neighbors classifier}}, where the number of neighbors (\texttt{n-neighbors}) is experimentally selected between 5 and 15. Although the \texttt{sklearn} library builds KNN models through various implementation strategies, like \texttt{BallTree}, \texttt{KDTree}, and \texttt{brute}, we opt for '\texttt{auto}', which helps us to decide the most appropriate implementation automatically depending on the input feeds.

\subsubsection{Naïve Bayes} \label{nb}
The naïve-Bayes classifier is built upon Bayes' theorem. The Bayes' theorem and conditional independence assumption of the predictor variables. The conditional independence assumption considers that the contribution of a predictor ($x$) on a given class ($c$) is independent of other predictors. 
In naïve-Bayes classifier, the probability of a classification label, $c_i$ for a given $m$-dimensional attribute vector, $A \in \{a_1, a_2, \cdots, a_m\}$ is computed as~\eqref{eq.Bayes-theorem}
\begin{equation}
    P(c_i|A) =\frac{P(A|c_i).P(c_i)}{P(A)},
    \label{eq.Bayes-theorem}
\end{equation}
where $P(A|c_i)$, $P(c_i)$, and $P(A)$ are the likelihood, prior, and marginal likelihood probabilities, respectively.  
Through conditional independence assumption and further simplification, \eqref{eq.Bayes-theorem} is estimated as \eqref{eq.n-Bayes}.
\begin{equation}
    P(c_i|A) = P(A)\prod_{j=1}^{m} P(a_j|c_i),
    \label{eq.n-Bayes}
\end{equation}
  
In this study, we use {sklearn's} \texttt{MultinomialNB} and \texttt{CategoricalNB} libraries\footnote[10]{\href{https://scikit-learn.org/stable/modules/generated/sklearn.naive_bayes.MultinomialNB.html}{scikit-learn NB classifier}} to built naïve Bayes models, where we set the parameter \texttt{fit-prior} to \texttt{True}, and  \texttt{class-prior} to \texttt{None}, so that the models learn class prior probabilities, since the datasets do not have a uniform class distribution among the samples. Hence, the smoothing parameter, \texttt{alpha} is optimally selected in the range of $[0, 1]$ through cross-validation methodology to overcome under-fitting and overfitting issues.

\subsubsection{Support Vector Machine}\label{svm}

SVMs identify a hyperplane or set of hyperplanes in high-dimensional space for classifying given set of samples. Kernel functions, such as polynomial, Gaussian and sigmoid are used for efficiently training SVM classifiers. For a given training data: $(x_1, y_1), (x_2, y_2), \cdots, (x_k, y_k)$, where input $x_i \in \mathbf{R}^n$ is the $i^{th}$ sample represented by an $n$-dimensional attributes, $y_i$ is the classification label of the input, and $k$ is the total number of samples. Then the class label of a testing data point, $x$ can be predicted from its probability density function, $p_i(x)$ using \eqref{eq.svm-hyperplane}.
\begin{equation}
    y = \arg \max p_i(x),
    \label{eq.svm-hyperplane}
\end{equation}
where $i=1, 2, \cdots, C$, $C$ being the total number of classes, and $p^i(x)=\sum_{j=1}^{k} y_i\lambda^i_j f(x, x_j) + b^i$, a signed real-valued logit, say, a belief score that can be considered as the distance from the decision boundary (separating hyperplane) to the test sample, $\lambda$ is a coefficient, $f(\cdot)$ is a penalty function, and $b$ is the bias. In this work, the SVM classifier for CVD prognosis is implemented using
the {sklearn's SVM} library\footnote[11]{\href{https://scikit-learn.org/stable/modules/generated/sklearn.svm.SVC.html}{scikit-learn support vector classifier}}, where the \texttt{Radial-basis function} (\texttt{rbf}) and linear kernel are selected for kernel parameter with squared $\ell_2$ penalty. Through empirical analysis, the final SVM classifier is configured with linear kernel achieving the best performance, because the data samples are found to be linearly separable.

\subsubsection{Evaluation Metrics} 

To validate the proposed CVD prognosis pipeline and to perform an exhaustive comparative analysis, the following standard evaluation metrics: accuracy~\eqref{eq.accuracy}, precision~\eqref{eq.precision}, and recall~\eqref{eq.recall} are used.
\begin{equation}
    Accuracy =\frac{TP+TN}{TP+FP+FN+TN},
    \label{eq.accuracy}
\end{equation}
\begin{equation}
    Precision =\frac{TP}{TP+FP},
    \label{eq.precision}
\end{equation}
\begin{equation}
    Recall =\frac{TP}{TP+FN},
    \label{eq.recall}
\end{equation}
where TP, TN, FP, and FN represent the true positive, true negative, false positive, and false negative, respectively. These quantities can be computed using a confusion matrix as shown in Fig.~\ref{conf_mat}.

\begin{figure}[!t]
\setlength\tabcolsep{4.5pt}    
\begin{tabular}{l|l|c|c|c}
\multicolumn{2}{c}{}&\multicolumn{2}{c}{True Prognosis}&\\
\cline{3-4}
\multicolumn{2}{c|}{} & \textcolor{red}{Positive} & \textcolor{green}{Negative} & \multicolumn{1}{c}{\textbf{Total}}\\
\cline{2-4}
\multirow{2}{*}{Prediction}&  \textcolor{red}{Positive} & $TP$ & $FP$ & $TP+FP$\\
\cline{2-4} & \textcolor{green}{Negative} & $FN$ & $TN$ & $FN+TN$\\
\cline{2-4}
\multicolumn{1}{c}{} & \multicolumn{1}{c}{\textbf{Total}} & \multicolumn{1}{c}{$TP+FN$} & \multicolumn{    1}{c}{$FP+TN$} & \\
\end{tabular}
    \caption{Illustration of A Confusion Matrix.}
    \label{conf_mat}
\end{figure}

We also measure performances of the models using area under the curve (AUC) of the receiver operating characteristic (ROC) curve, where ROC is a probability curve and AUC represents the degree or measure of separability. The ROC curve is plotted with recall~\eqref{eq.recall} against the specificity~\eqref{eq.specificity}, where recall and specificity are represented on y-axis and x-axis, respectively (cf.~Fig.\ref{AUC-ROC-plot-all-data-set}).
\begin{equation}
    Specificity =\frac{FP}{TN+FP},
    \label{eq.specificity}
\end{equation}

\setlength\tabcolsep{2pt}
\begin{table}[!t]
\caption{Summary of The Hyperparameter Settings. Beyond These List Of Hyperparameter Settings, All Other Hyperparameter Values Are Left with Their Default Configurations}
    \label{tab:Hyperparameters}
    \centering
    \begin{tabular}{|p{2.5cm}|p{2.8cm}|p{2.9cm}|}
    \hline\hline
        \textbf{Classifier} &  \textbf{Hyperparameter} & \textbf{Hyperparameter Value} \\
        \hline\hline
        Logistic regression & {solver}  & {newton-cg}  \\\hline
        
        \vspace{0.005cm}Decision tree & {criterion}, {splitter}, {max-depth} &  {gini/entropy}, {best}, {3 - 6} \\\hline
        
        \vspace{0.005cm}K-NN & {n-neighbors}, {algorithm}, {metric} & {5 - 15}, {auto}, {euclidean}
        \\\hline
        
        Multinomial NB &  {alpha} & {0.5 - 0.8}  \\\hline
        
        Categorical NB & {alpha} & {0.5 - 0.8}  \\\hline
        
        SVM&  {kernel} & {rbf/linear} \\\hline\hline
        
    \end{tabular}
    \footnotesize{Note: The names of the hyperparameters are given based on Scikit-learn Python libraries - https://scikit-learn.org/stable/.}
    
\end{table}

\subsection{Model Deployment}\label{model-deployment}
The best performant model found through investigation in this work is deployed on a web-based system for end-user application, which will help users to take precautionary measures regarding heart diseases (note that the website will be publicly available after the acceptance of this paper). It will also help us collect more samples for refining the proposed solution, and for future research and development supporting the healthcare infrastructure in Bangladesh and elsewhere.

\begin{algorithm}[!t]
  \caption{An Algorithmic Summary of the Proposed Cardiovascular Diseases Prognosis Pipeline}
  \label{algo-summary}
    $\textbf{Require:}$ Datasets - BHDC, Cleveland, Hungarian, Switzerland, and Long Beach VA datasets. \\
    $\textbf{Import:}$ Pandas, sklearn, numpy, time, and other required libraries.\\
    
    \KwData{read datasets using Pandas builtin function}
    \textbf{Pre-processing:}{ Data cleaning and data transformation.\\
        \eIf{BHDC dataset}{
        Follow the steps given in Section~\ref{sec-BHDC}\;
        }{
        Follow the steps given in Section~\ref{sec-ucis'}\;
        
        }
    }
    \textbf{Shuffle:} Sample randomization using Python builtin function\;
    \textbf{Feature selection method initialization:}\\
            $\beta \xleftarrow{}$ Expert recommendation\;
            $\alpha \xleftarrow{}$ ANOVA-based feature selection\;
            $\eta \xleftarrow{}$ Fusion of $\beta$ and $\alpha$\;
    \textbf{Examination:} K-fold cross-validation\;
    \If{$\beta$ or $\eta$}{
        Build CVD prognosis models using the classifiers stated in Sections~\ref{lr}, \ref{dt}, \ref{knn}, \ref{nb}, and \ref{svm}\;
        Validate the models' performances using the evaluation metrics stated in \eqref{eq.accuracy}, \eqref{eq.precision}, \eqref{eq.recall}, AUC, and inference speed\;
    }
    \If{$\alpha$}{
            Set $n = 2$ \textcolor{gray}{// top-n feature selection via ANOVA}\;
            \While{$n<= 18$}{
                $\alpha \xleftarrow{}$ ANOVA-based top-$n$ attributes\;
                Build CVD prognosis models using the classifiers stated in Sections~\ref{lr}, \ref{dt}, \ref{knn}, \ref{nb}, and \ref{svm}\;
                Validate the models' performances using the evaluation metrics stated in \eqref{eq.accuracy}, \eqref{eq.precision}, \eqref{eq.recall}, AUC, and inference speed\;
                $n = n + 2$\;
            }
    }
    
    \If{$\eta$}{
        $\alpha \xleftarrow{}$ The best performant top-$n$ feature sets from ANOVA\;
        $\eta \xleftarrow{}$ $Concat(\alpha, \beta)$ \;
        Build CVD prognosis models using the classifiers stated in Sections~\ref{lr}, \ref{dt}, \ref{knn}, \ref{nb}, and \ref{svm}\;
        Validate the models' performances using the evaluation metrics stated in \eqref{eq.accuracy}, \eqref{eq.precision}, \eqref{eq.recall}, AUC, and inference speed\;
    }
    
  \end{algorithm}

\section{Experimental Study and Analysis}
\label{sec:experiments}

\subsection{Dataset Summary}\label{datasets}
This study uses five datasets, including its own collection, the BHDC, as well as, four benchmark datasets from the UCI machine learning repository to validate the proposed pipeline for heart disease prognosis. Table~\ref{tab:datasets_summary} summarizes the above mentioned datasets and their attribute details. For information about data curation, please refer to Section~\ref{data-curation}. 

\subsection{Experimental Setup}\label{Experimental}

The experimental setup of the proposed model is summarized in Algorithm~\ref{algo-summary}, in which the model initialization and hyperparameter tuning are the core components. 

\subsubsection{Model Initialization and Hyperparameter Tuning}

The key details of various model initialization and their respective hyperparameter settings are elaborated in subsections~\ref{lr}, \ref{dt}, \ref{knn}, \ref{nb}, and \ref{svm}. Hence, Table~\ref{tab:Hyperparameters} summarizes the important hyperparameters and their values used in this work with respective to the classifiers.

\subsubsection{Software Paradigm} 

This work uses Python Ver. 3.8.8 and its open-source ML libraries on a Windows 10 64-bit OS for model building, training, and validation of the proposed CVD prognosis pipeline. For deployment of the proposed pipeline, a web-based user interface is developed using HTML, CSS, Bootstrap, and FastAPI.


\subsubsection{Hardware Paradigm} 

The proposed pipeline for CVD prognosis is developed and tested on a computational platform having an AMD A6-6310 APU, and 8.00 GB memory. 

\subsection{Step-by-Step Quantitative Analysis}\label{Quantitative-Analysis}

The quantitative ablation study is presented in a step-by-step manner that includes overall and database-specific comparative analysis wrt state-of-the-art models. 

\begin{figure*}[!ht]
    \centering
    \footnotesize{\hspace{1cm} BHDC \hspace{2.5cm} Cleveland \hspace{2.0cm} Switzerland \hspace{2.1cm} Hungarian \hspace{2.0cm} Long-beach-Va}\vspace{-0.3cm} \newline
    \includegraphics[trim={0.5cm, 0.9cm, 0.6cm, 2.0cm}, clip, width=.19\textwidth]{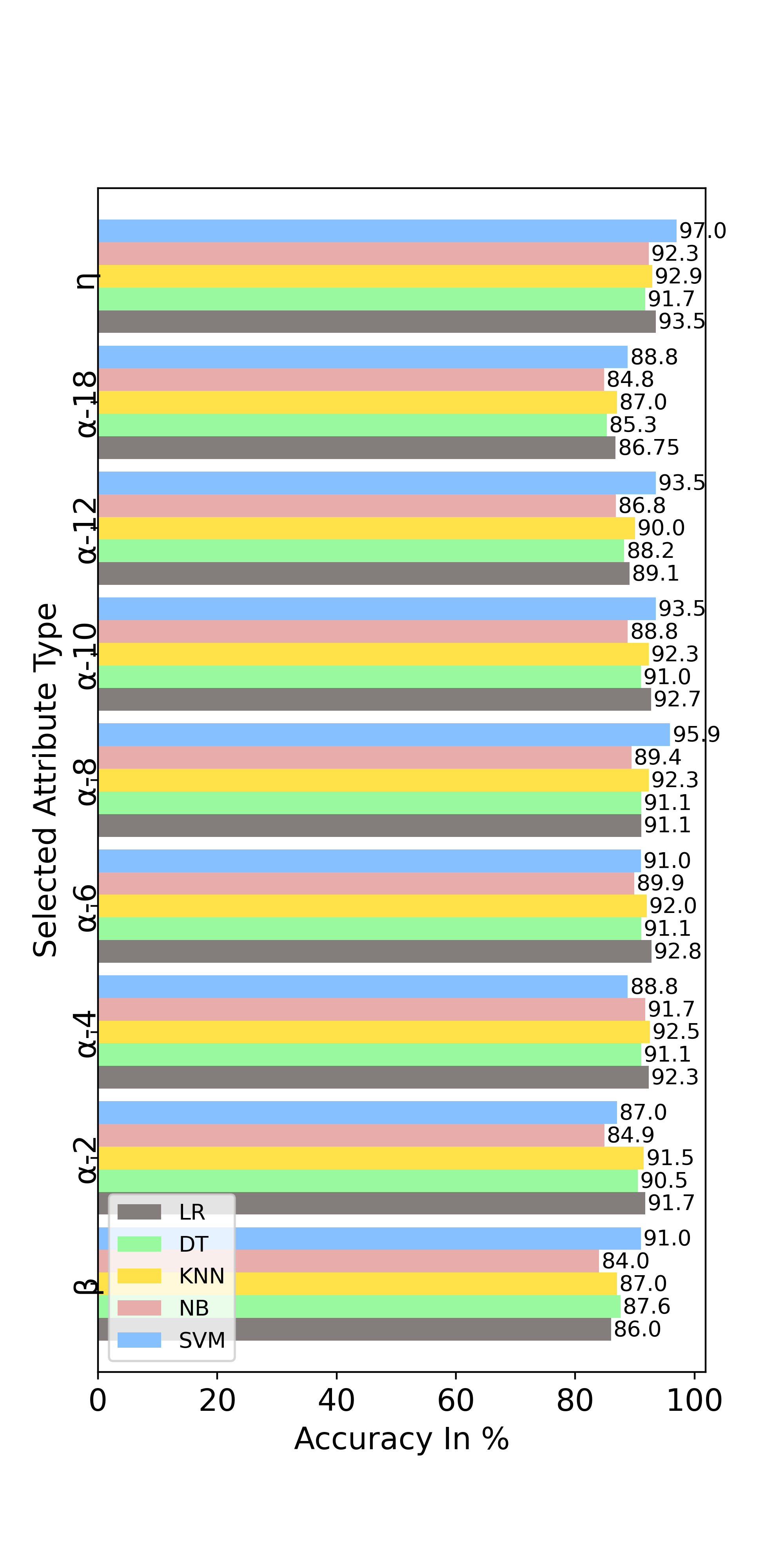}
    \includegraphics[trim={0.5cm, 0.9cm, 0.6cm, 2.0cm}, clip, width=.19\textwidth]{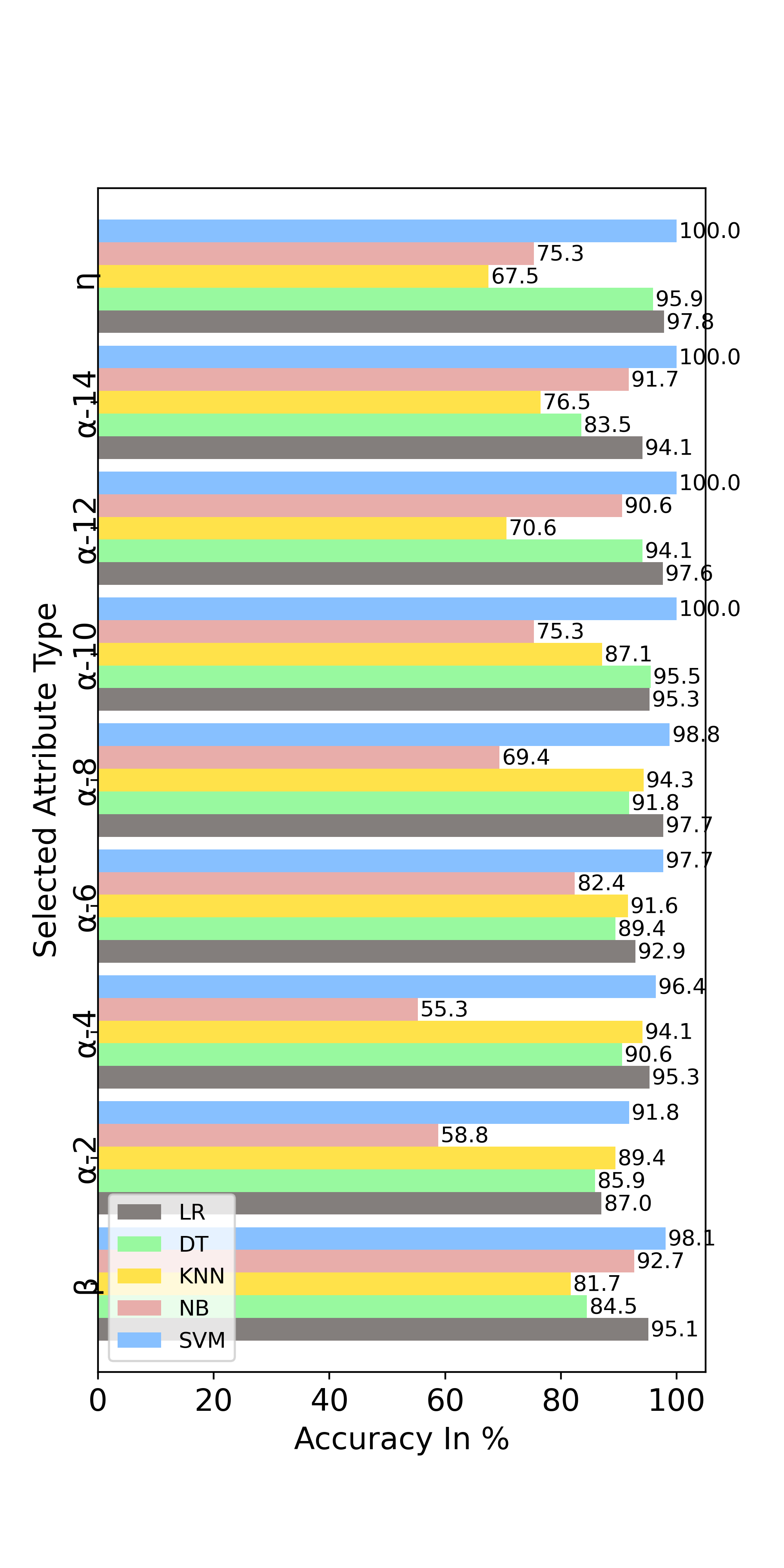}
    \includegraphics[trim={0.5cm, 0.9cm, 0.6cm, 2.0cm}, clip, width=.19\textwidth]{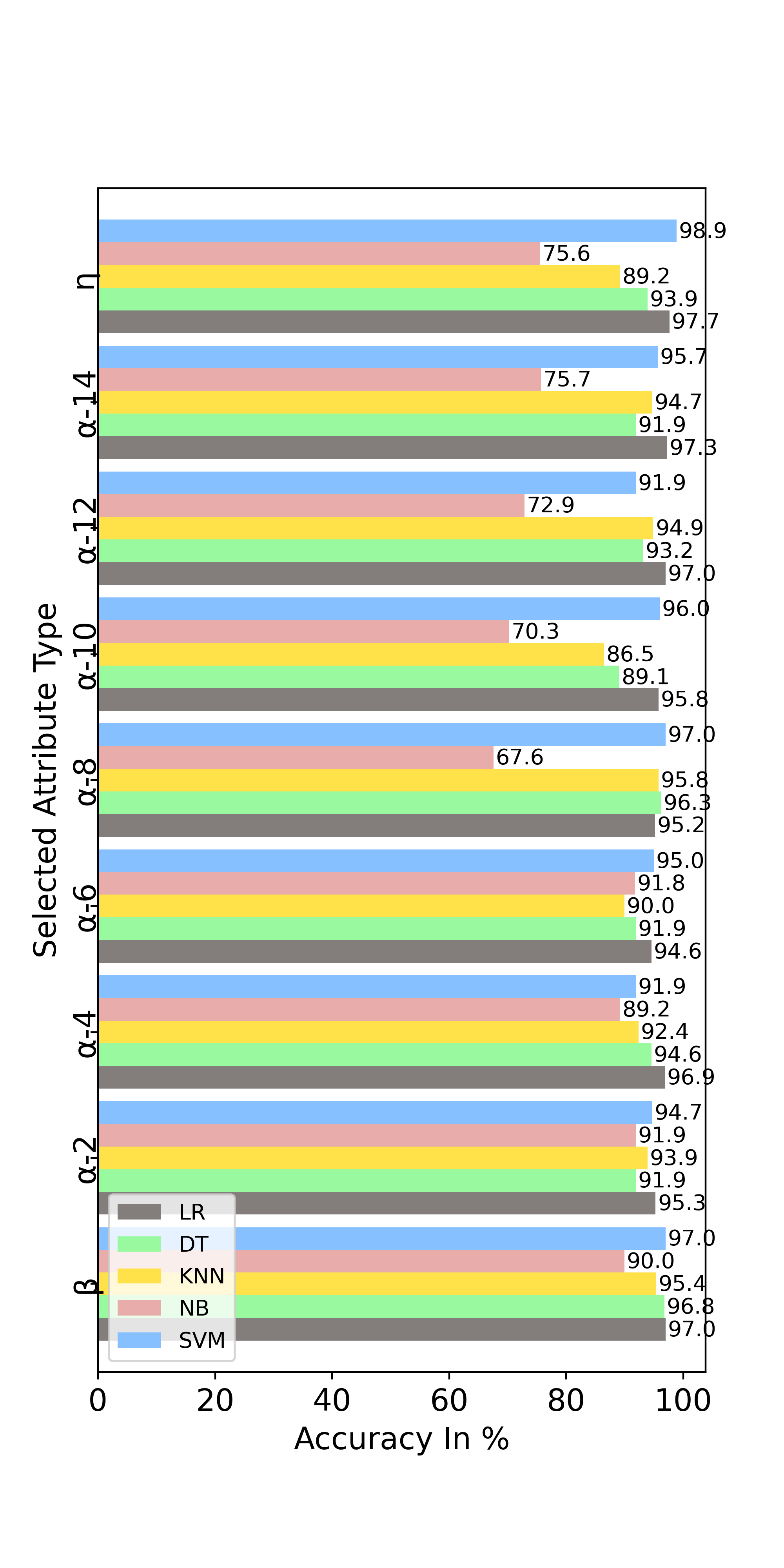}
    \includegraphics[trim={0.5cm, 0.9cm, 0.6cm, 2.0cm}, clip, width=.19\textwidth]{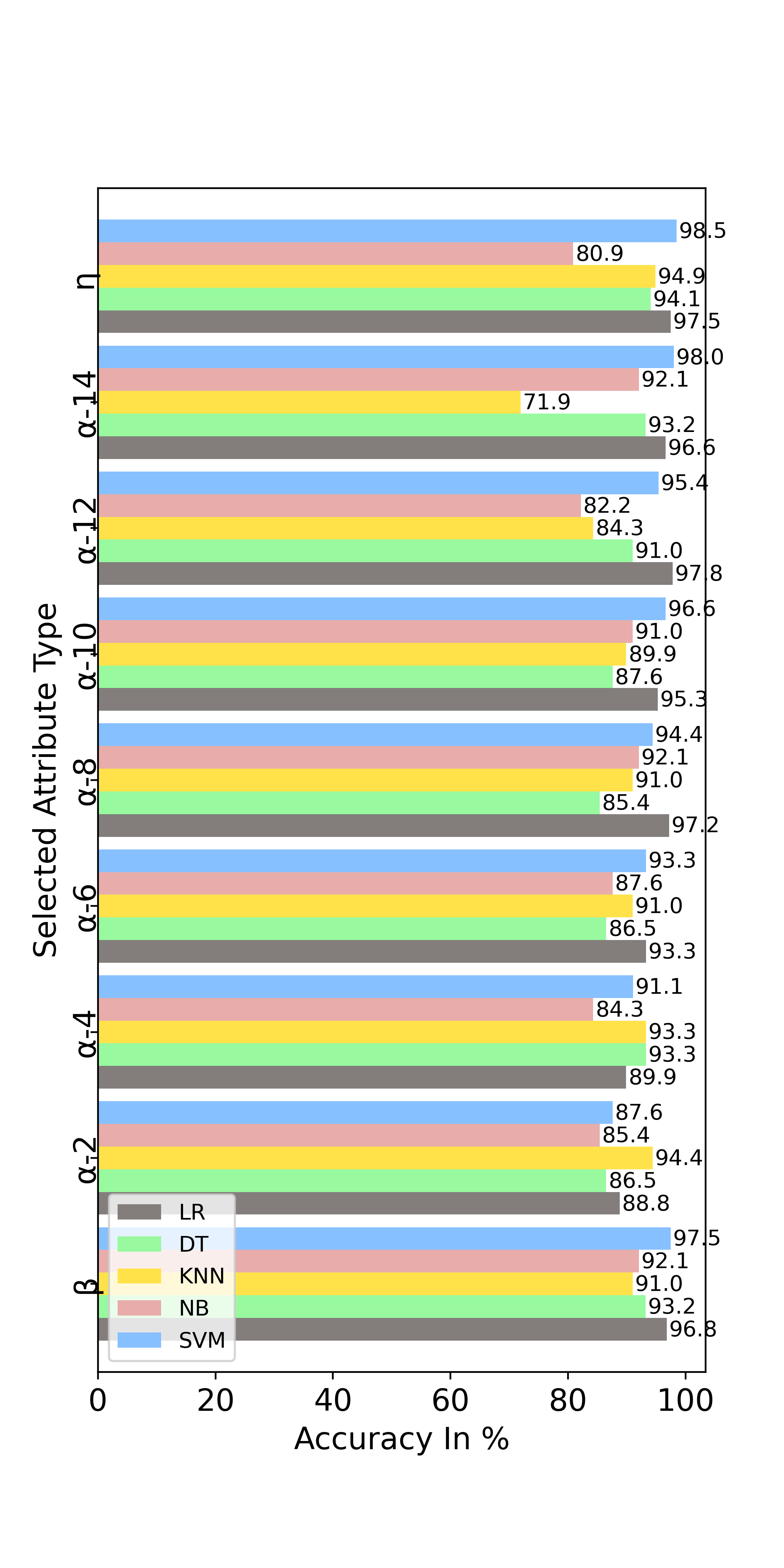}
    \includegraphics[trim={0.5cm, 0.9cm, 0.6cm, 2.0cm}, clip, width=.19\textwidth]{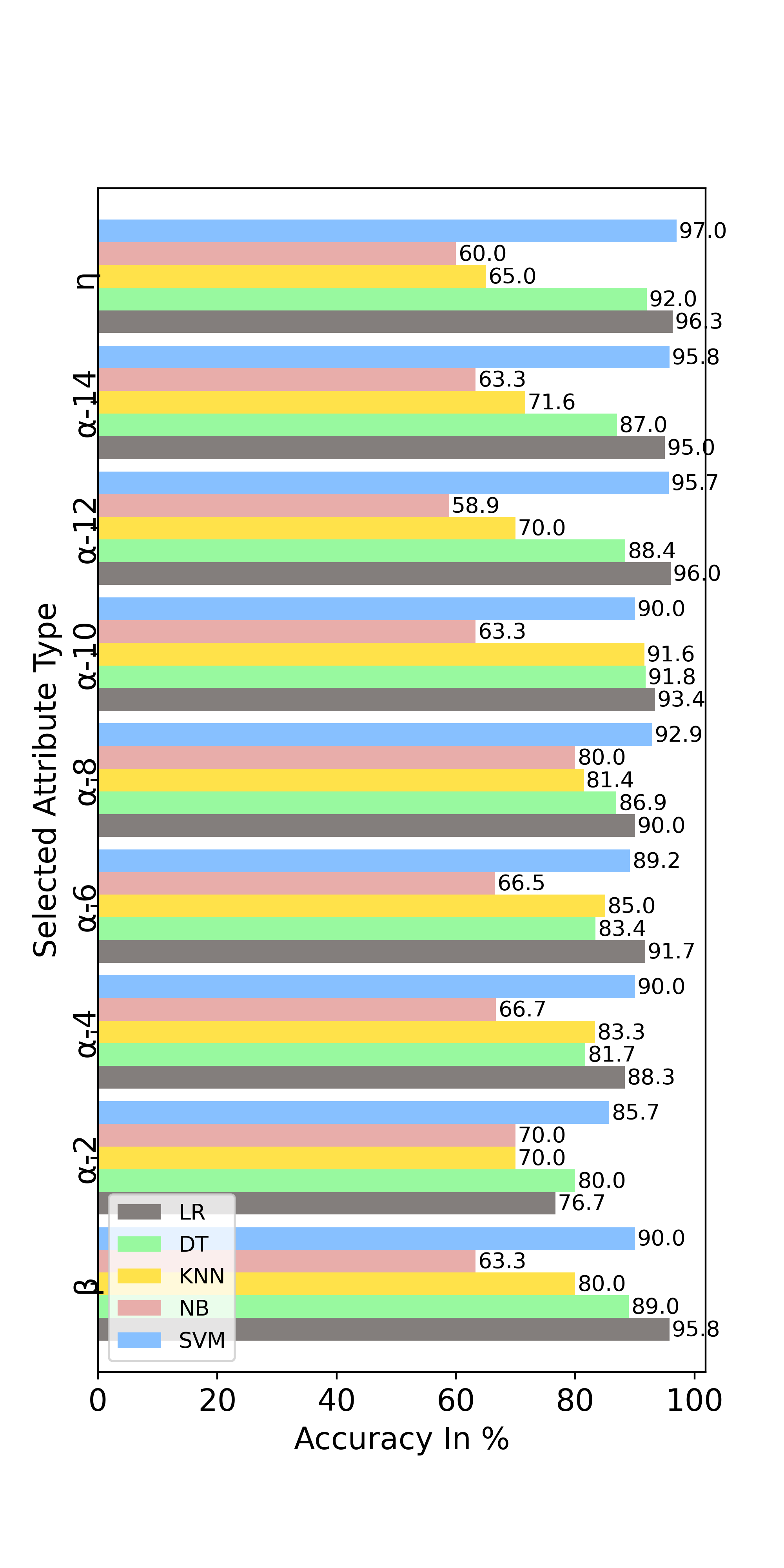}
    \caption{Performance of the Various Models wrt the Type of Attribute Selection: $\beta$ - Expert recommendation, $\alpha$ - ANOVA-based top-n feature selection, and $\eta$ - Fusion of $\beta$ and $\alpha$ on Five Different Datasets. }
    \label{performance-of-all-models}
\end{figure*}

\begin{figure*}[!ht]
\centerline{\includegraphics[trim={0.8cm, 0.6cm, 1.5cm, 1.7cm}, clip, width=0.48\textwidth]{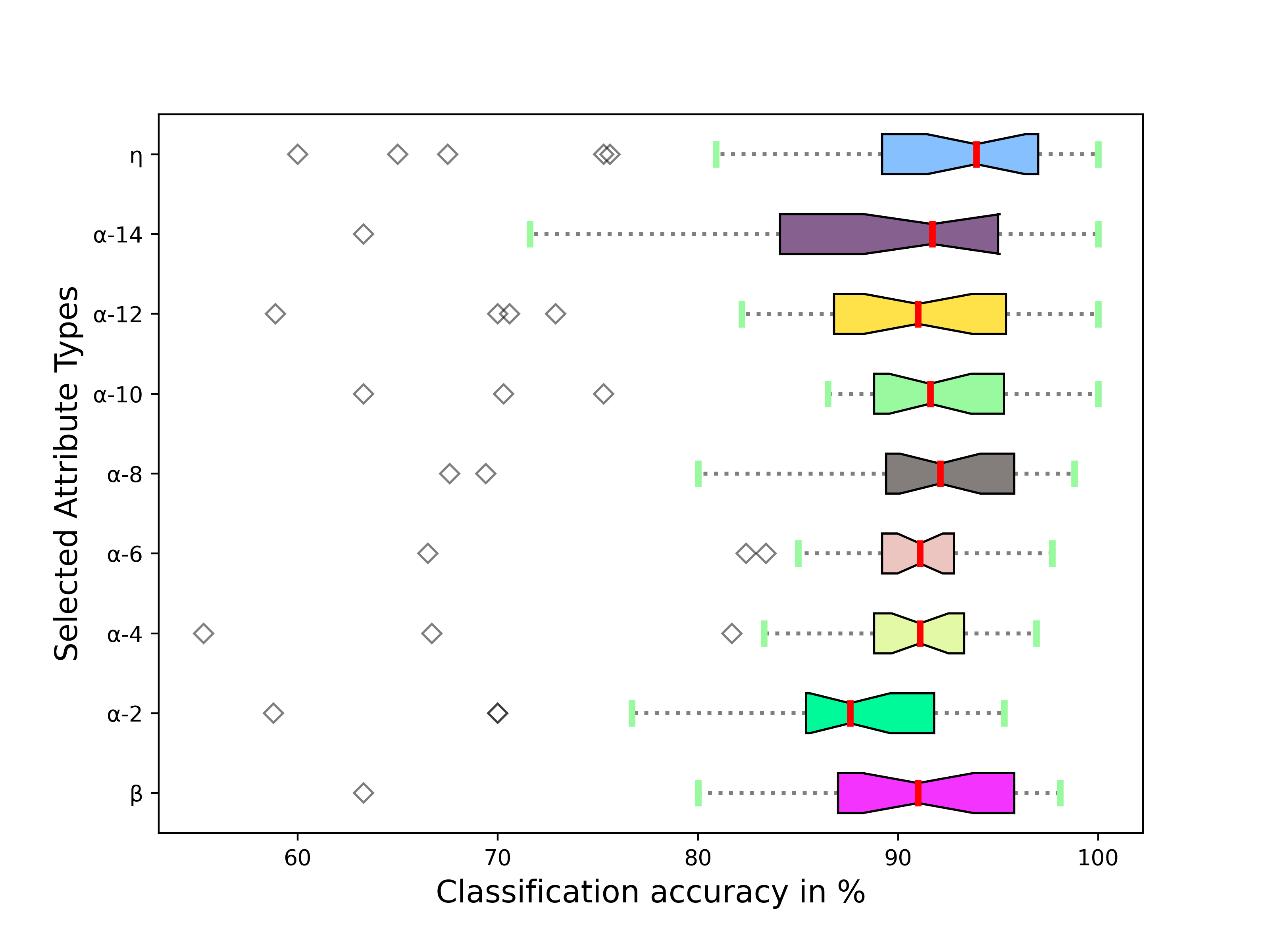}
\includegraphics[trim={0.8cm, 0.6cm, 1.5cm, 1.7cm}, clip, width=0.48\textwidth]{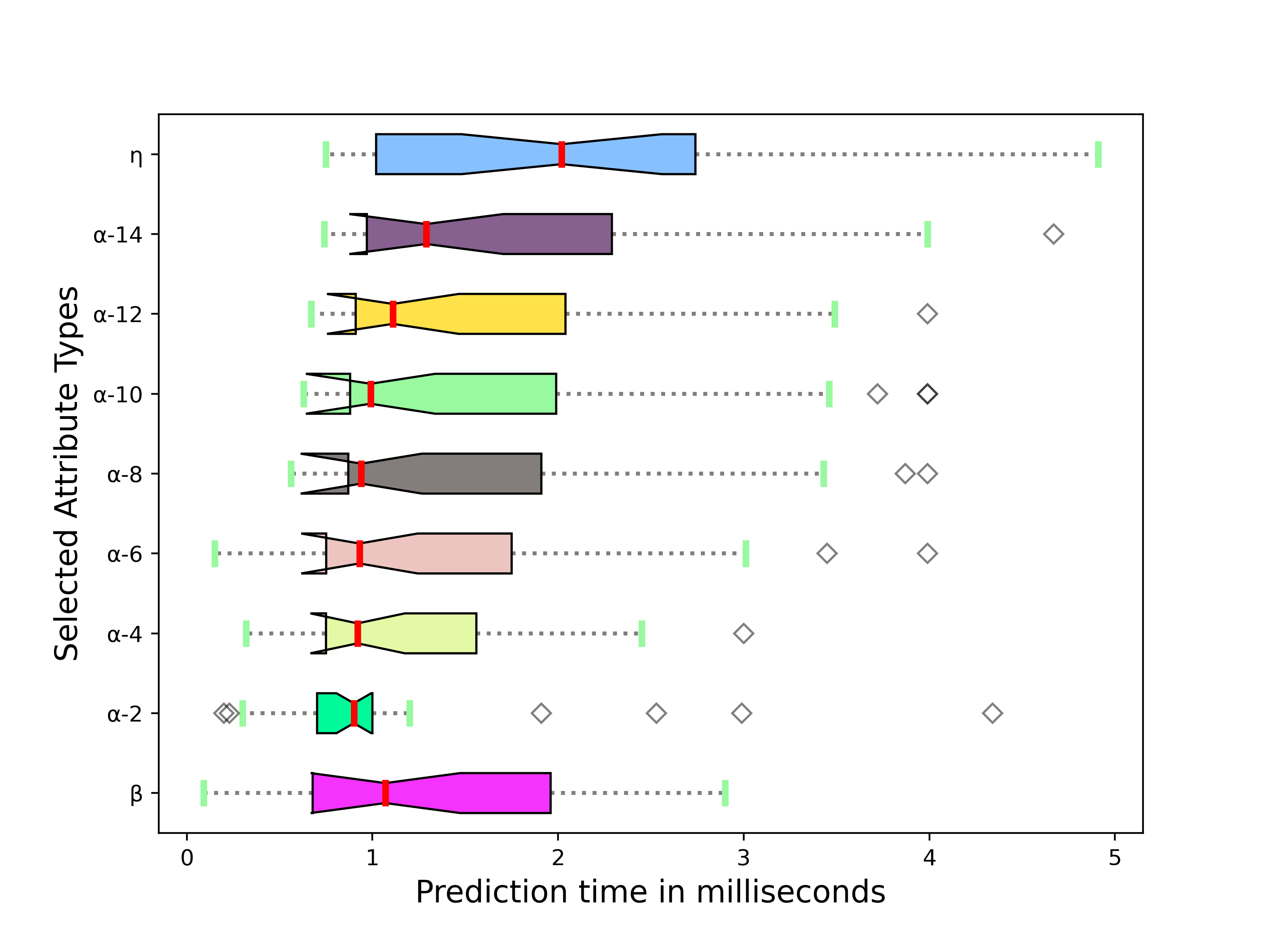}
}
\caption{Robustness Analysis of Feature Selection Approaches. Col. 1 and Col. 2 Summarize Nine Different Feature Selection Approaches' Classification Performances and Per Sample Prediction Times Across All the Classifiers and Datasets, Respectively.}
\label{feature-selection-with-accuracies-time}
\end{figure*}

\begin{figure*}[!ht]
\centerline{
\includegraphics[trim={0.6cm, 0cm, 1.5cm, 1.0cm}, clip, width=0.475\textwidth]{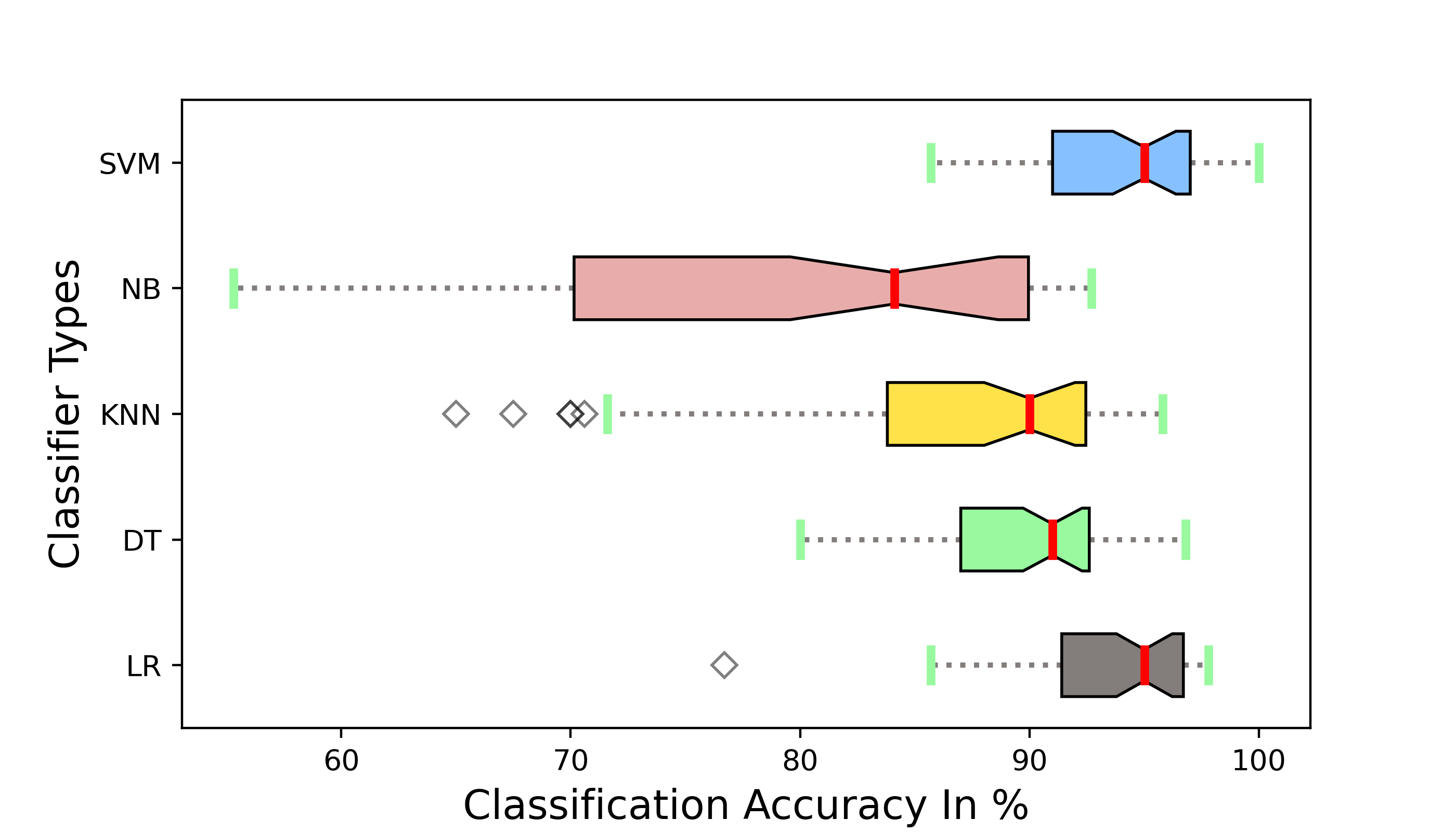}
\includegraphics[trim={0.6cm, 0.1cm, 1.5cm, 1.0cm}, clip, width=0.475\textwidth]{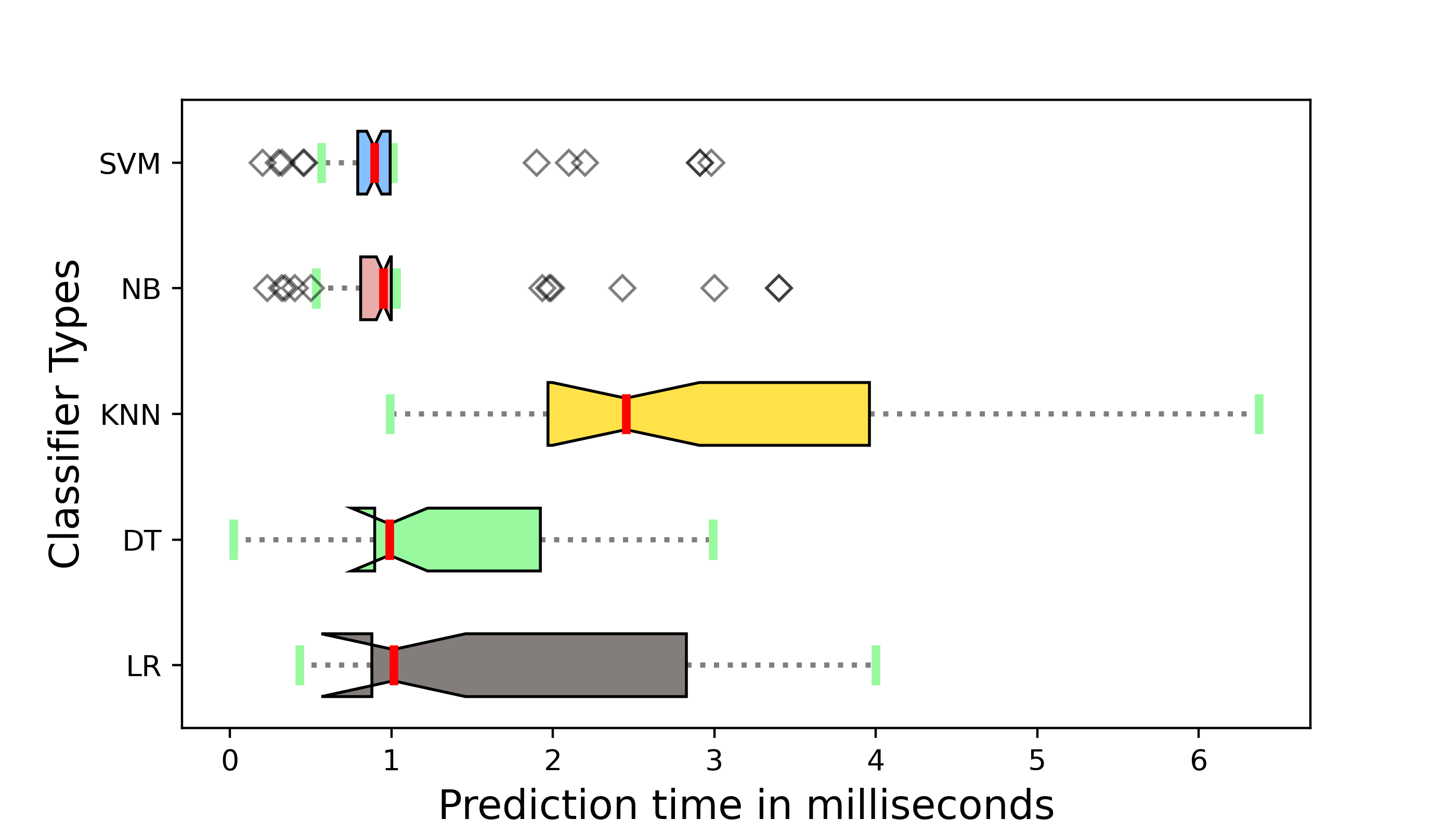}
}
\caption{Robustness Analysis of the Classifiers. Col. 1 and Col. 2, Respectively, Summarize Five Different Classifiers' Classification Accuracies and Per Sample Prediction Time Across All the Types of Attribute Selections and Datasets.}
\label{models-accuracies-and-time}
\end{figure*}

\begin{figure*}[!ht]

\footnotesize{\hspace{1.8cm} BHDC \hspace{3.5cm} Cleveland \hspace{3.2cm} Switzerland \hspace{3.2cm} Hungarian }\vspace{-0.2cm} \newline
\centerline{
\includegraphics[trim={0.2cm, 0.0cm, 0.9cm, 0.9cm}, clip, width=0.25\textwidth]{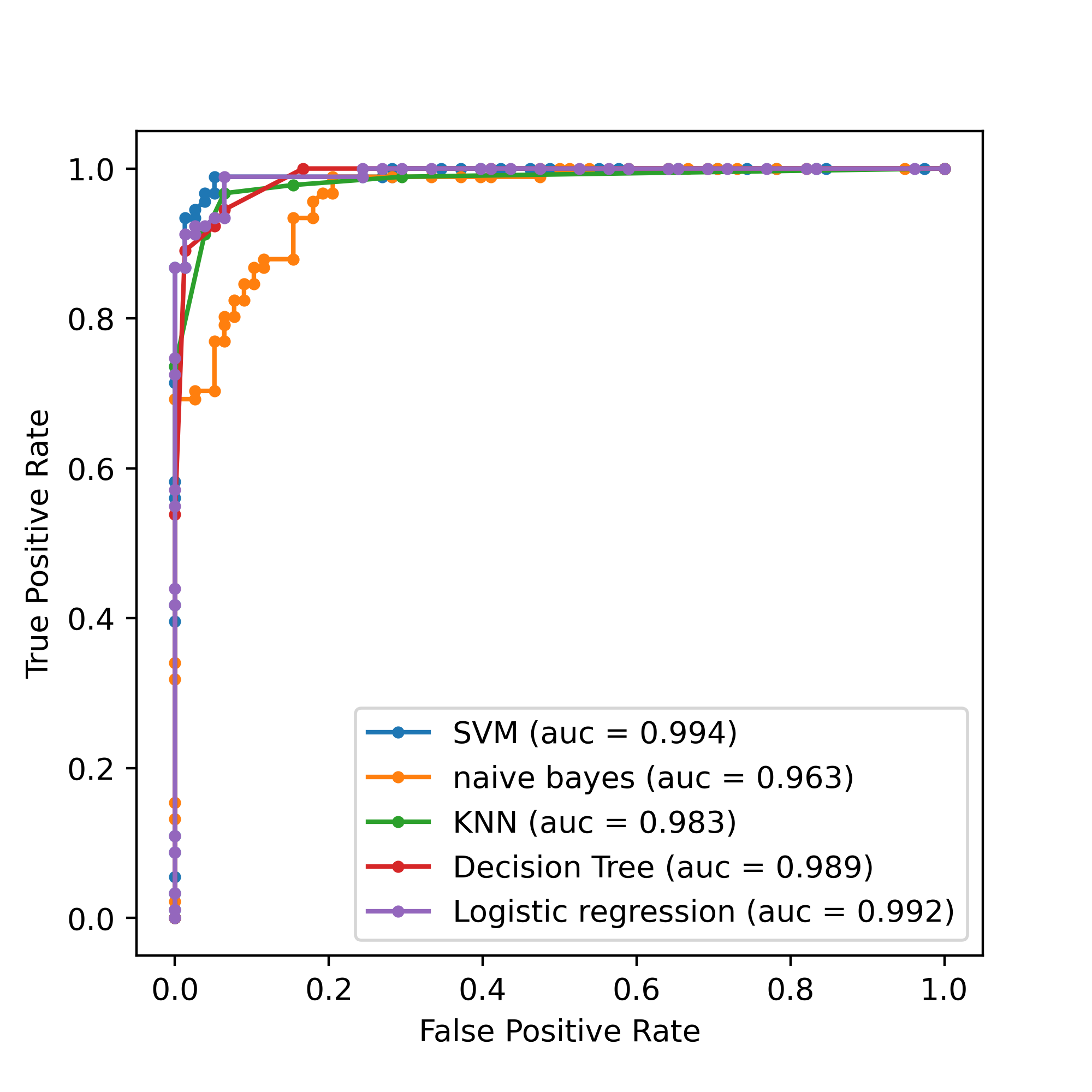}
\includegraphics[trim={0.2cm, 0.0cm, 0.9cm, 0.9cm}, clip, width=0.25\textwidth]{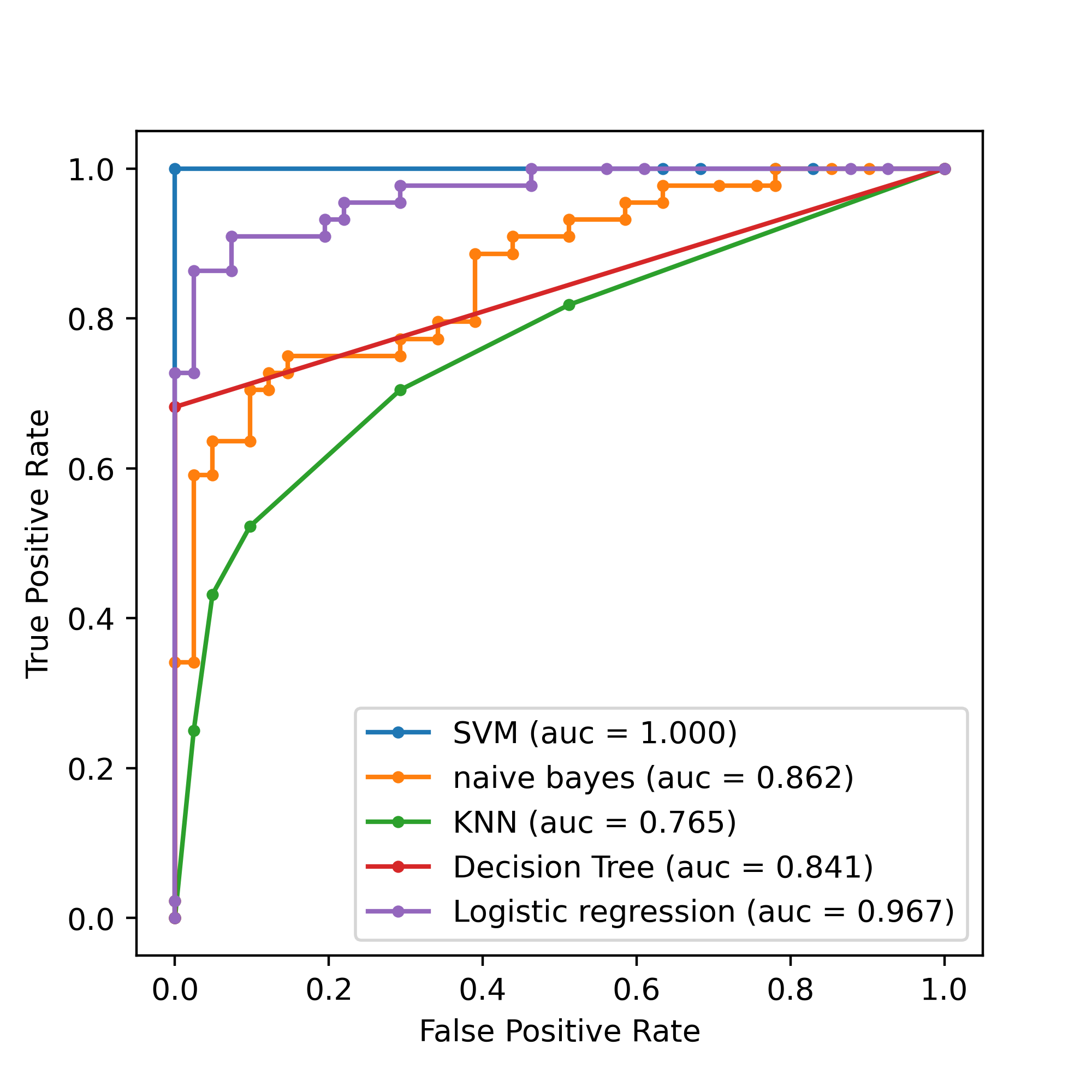}
\includegraphics[trim={0.2cm, 0.0cm, 0.9cm, 0.9cm}, clip, width=0.25\textwidth]{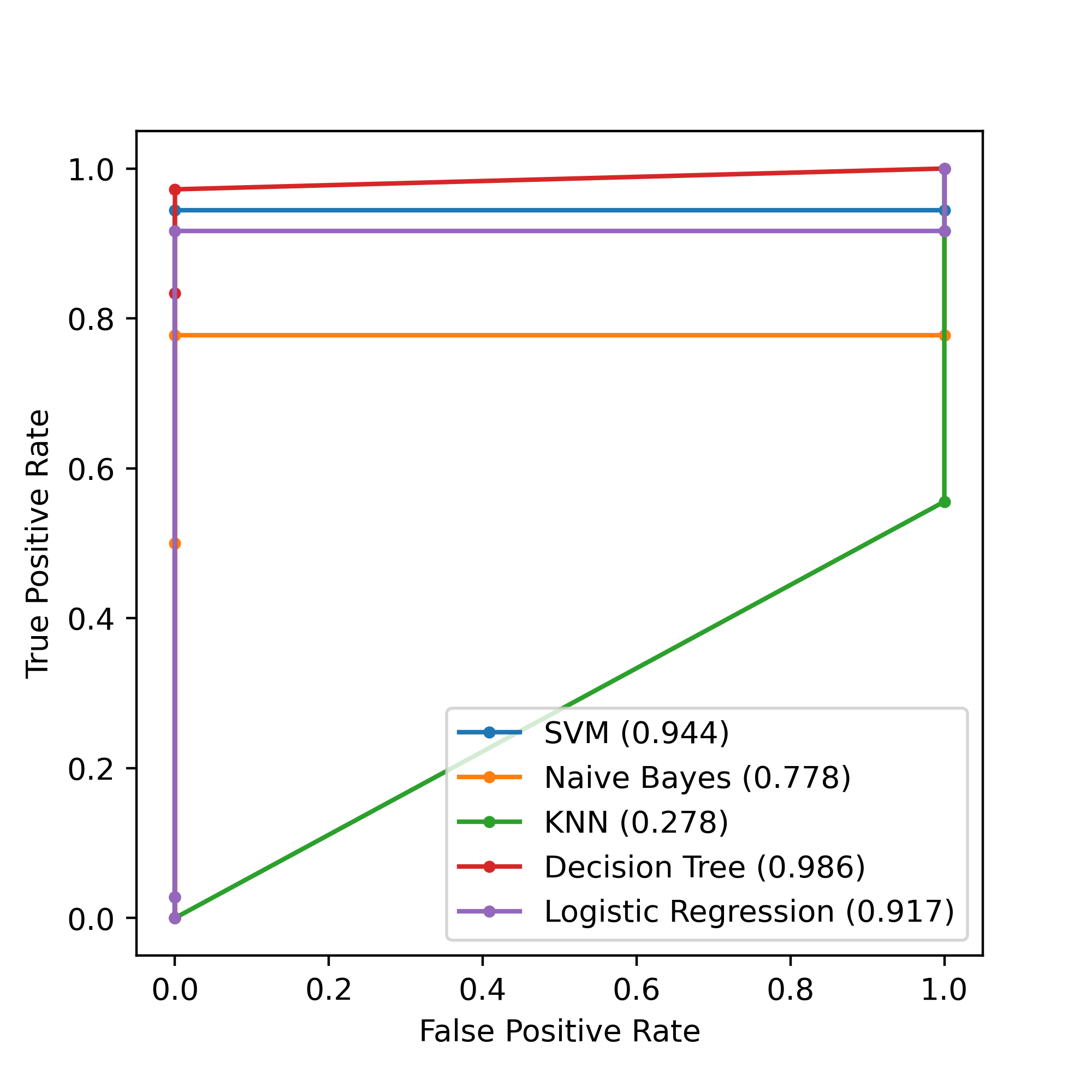}
\includegraphics[trim={0.2cm, 0.0cm, 0.9cm, 0.9cm}, clip, width=0.25\textwidth]{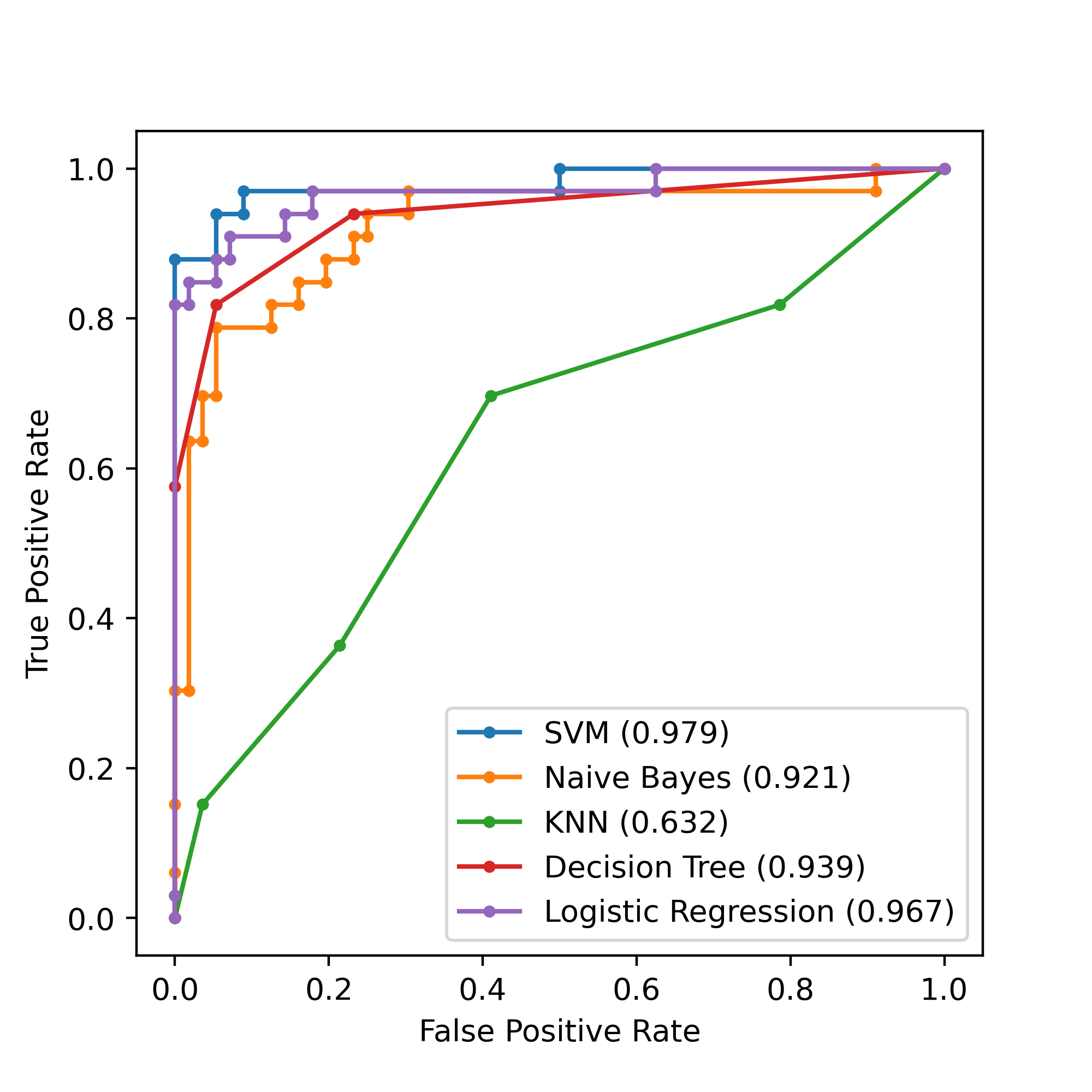}
}
\caption{AUC-based Performance Analysis of Various Classifiers using Fusion-based Attribute Selection. Col. 1 - 4 Show the ROC of the Models When Tested on BHDC, Cleveland, Switzerland, and Hungarian Datasets.}
\label{AUC-ROC-plot-all-data-set}
\end{figure*}

\begin{table*}[!htp]
\begin{center}
\caption{Overall Performance Analysis: the Best Results Achieved in this Work vs. the Existing State-of-the-Art Models. $'-'$ Denotes that the Existing Work Did Not Use the Specific Dataset. Note that the Baseline Model's Performance is Underlined.} 
\label{tab-result-comparison}
\small
\setlength\tabcolsep{4pt}

\begin{tabular}{|m{2.0cm}|m{0.8cm}|m{1.7cm}|m{1.5cm}|m{2.3cm}|m{2.0cm}|m{1.8cm}|m{1.8cm}|m{1.5cm}|}
\hline \hline
\multirow{2}[8]{*}{\textbf{Dataset}} & \multicolumn{8}{c|}{\textbf{Various Model's Performances in Percentage (\%)}}\\ \cline{2-9}
& This Work & \centering Mohan~\textit{et~al.} \cite{mohan2019effective} & \centering Shah~\textit{et~al.} \cite{shah2020heart}  & \centering G{\'a}rate-Escamila~\textit{et~al.} \cite{garate2020classification} &  \centering El-Bialy~\textit{et~al.} \cite{el2015feature} & \centering Babu~\textit{et~al.} \cite{babu2018medical} &  \centering  Anooj~\textit{et~al.} \cite{anooj2012clinical} & {\centering
Uyar~\textit{et~al.} \hfill \quad \hspace{3cm} \cite{UYAR2017588} } \\ 
\hline 
\hline


Cleveland & \centering \textcolor{blue}{$100.0$}  &  \centering $88.7$ & \centering $90.8$ & \centering \underline{$98.7$} & \centering $78.5$ & \centering  $98.2$ & \centering $76.5$ &  \quad\quad $97.8$ \\
\hline

Hungarian & \centering \textcolor{blue}{$99.4$}  & \centering $-$ & \centering $-$ & \centering \underline{$99.0$} & \centering $78.5$ & \centering $95.1$ & \centering $74.2$ & \quad\quad $-$\\
\hline

Switzerland & \centering \textcolor{blue}{$99.0$}  & \centering $-$ & \centering $-$ & \centering $-$ & \centering $-$ & \centering \underline{$91.3$} & \centering $52.6$ &  \quad\quad $-$\\
\hline

Long Beach Va & \centering \textcolor{blue}{$98.4$}  & \centering $-$ & \centering $-$ & \centering $-$ & \centering \underline{$71.5$} & \centering $-$ & \centering $-$ & \quad\quad $-$\\ 
\hline\hline


\end{tabular}
\end{center}

\end{table*}

\subsubsection{Overall Analysis}

\textbf{Best feature selection method -} Fig.~\ref{performance-of-all-models} demonstrates the impact of various feature selection approaches in terms of classifiers' performances on all benchmark datasets. 
It is found that among all the feature selection methods, the proposed fusion-based approach ($\eta$) achieves the best results in all the benchmark datasets. To analyse the robustness of various feature selection strategies, Fig.~\ref{feature-selection-with-accuracies-time} visualizes the performance of each feature selection approach wrt classification accuracies and per sample prediction times considering all types of classifiers and datasets. This analysis further confirms that the feature fusion method $\eta$ is a robust feature selection strategy for CVD prognosis, although there is a negligible overhead of processing time.

\textbf{Best classifier selection-} Through summarizing the performances of all the models, considering all feature selection techniques and datasets via box plots as shown in Fig.~\ref{models-accuracies-and-time}, it is evident that the model with SVM outperforms all other classifiers without any compromising in the prediction time. To further validate proposed fusion-based CVD prognosis model, AUC of all the models are computed as shown in Fig.~\ref{AUC-ROC-plot-all-data-set}. This performance analysis also proves that the model with SVM and fusion-based automatic attribute selection ($\eta$) consistently provides the best results across all the benchmark datasets.

\textbf{Comparison to the existing works -} Table~\ref{tab-result-comparison} compares the best performant model presented in this work, i.e., the SVM classifier integrated with the proposed ANOVA-based automatic feature selection and fusion with the state-of-the-art existing approaches. 
It is unfortunate that not all the existing works report a complete experimental study across all the benchmark datasets. 
However, we have carried out a thorough experimental analysis on all the datasets, and our own dataset -- BHDC, which contains more data samples than any of the benchmark datasets (cf.~Table~\ref{tab:datasets_summary}). 
When compared to the baseline models' performances (highlighted by underscores in Table~\ref{tab-result-comparison}), the proposed model improves the accuracy of the CVD prognosis by $1.3\%$, $0.4\%$, $7.7\%$, and $26.9\%$, respectively, on Cleveland, Hungarian, Switzerland, and Long Beach Va datasets. 
On the other hand, on the BHDC dataset our proposed model records the best classification accuracy of $98.6\%$ (cf.~Table~\ref{tab-result-BHDC}).

\begin{table*}[!t]
\begin{center}
\caption{Various Classifiers' Performance Analysis on BHDC Dataset wrt Type of Feature Selection. Note: $\beta$ - Expert Recommended Features, $\alpha$-$n$ - ANOVA-based Top-$n$ Features, $\eta$ - Fusion of $\beta$ and $\alpha$, LR - Logistic Regression, DT - Decision Tree, K-NN - K Nearest Neighbor, NB - Naïve Bayes, and SVM - Support Vector Machine.} 
\label{tab-result-BHDC}
\small
\setlength\tabcolsep{4pt}

\begin{tabular}{|p{1.7cm}|p{1.1cm}|p{1.1cm}|p{1.2cm}|p{1.2cm}|p{1.1cm}|p{1.2cm}|p{1.2cm}|p{1.2cm}|p{1.2cm}|p{1.2cm}|p{1.1cm}|}
\hline \hline
\multirow{2}[1]{*}{\textbf{Model Name}}  & \multicolumn{11}{c|}{\textbf{Accuracy (in $\%$) of CVD Prognosis Using Various Feature Selection Types}}\\ \cline{2-12}
& {$\beta$} & {$\alpha$-2}
& $\alpha$-4  & $\alpha$-6 & $\alpha$-8 & $\alpha$-10 & $\alpha$-12 & $\alpha$-14 & $\alpha$-16 & $\alpha$-18 & $\eta$ \\ 
\hline 
\hline

LR & 86.0±0.1 & 91.7±2.2 & 92.3±0.0 & 92.8±0.03 & 91.1±0.6 & 92.7±0.08 & 89.1±1.0 & 90.0±1.1 & 85.7±0.1 & 86.75±1.1 & \textcolor{blue}{93.5±0.4}\\
\hline

DT &87.6±0.3 &90.0±1.0 &91.1±0.01 &91.1±0.01 &91.1±0.2  &91.0±0.01 &88.2±1.0 &88.1±1.0  &87.0±0.3  &85.3±3.0  &\textcolor{blue}{91.7±0.2}\\
\hline

K-NN &87.0±0.2  &90.0±1.0 &92.5±0.2 &92.0±0.9 &92.3±0.1  &92.3±0.0 &90.0±2.0 &90.0±0.0  &87.0±0.3  &87.0±1.9 &\textcolor{blue}{92.9±0.1}\\
\hline

NB &84.0±0.6  &84.9±3.0 &91.7±0.1 &89.9±0.2 &89.4±0.3 &88.8±0.4 &86.8±0.7 &84.8±1.1  &84.1±0.1  &84.8±1.3 &\textcolor{blue}{92.3±0.2}\\
\hline

SVM &91.0±0.4 &87.0±3.1 &88.8±4.0 &91.0±0.2 &95.9±0.3 &93.5±1.1 &93.5±0.6 &90.0±0.8  &90.2±0.0  &88.8±0.9 &\textcolor{blue}{97.0±1.6}\\

\hline\hline
\end{tabular}
\end{center}
\end{table*}

\subsubsection{Performance Analysis on The BHDC Dataset}

Table~\ref{tab-result-BHDC} provides the quantitative performance analysis of eleven feature selection approaches applied on the BHDC dataset. These approaches are the expert recommended features ($\beta$), nine ANOVA-based automatic top-n features ($\alpha$-$n$), and the proposed fusion-based feature selection ($\eta$). Recall that this fusion approach concatenates the $\beta$ and the best performant ANOVA-based top-n features. In this case, $\alpha$-8 is the best performant ANOVA-based top-n features. 
This comparative analysis shows that the SVM model with the proposed feature fusion technique -- $\eta$ achieves the best classification results of $97.0$±$1.6\%$. Note that in Table~\ref{tab-result-BHDC}, the classifiers' best performances are inked in \textcolor{blue}{blue}.


\begin{table*}[!ht]
\begin{center}
\caption{Various Classifiers' Performance Analysis on UCI Datasets: Cleveland, Switzerland, Hungarian, and Long Beach Va Datasets. Note: $\beta$ - Expert Recommended Features, $\alpha$-$n$ - ANOVA-based Top-$n$ Features, $\eta$ - Fusion of $\beta$ and $\alpha$, LR - Logistic Regression, DT - Decision Tree, K-NN - K Nearest Neighbor, NB - Naïve Bayes, and SVM - Support Vector Machine.}
\label{tab-result-ucis-datasets}
\small
\setlength\tabcolsep{4pt}
\begin{tabular}{|l|l|c|c|c|c|c|c|c|c|c|}
\hline \hline
\multirow{2}[1]{*}{\textbf{Dataset Name}} &
\multirow{2}[1]{*}{\textbf{Model Name}}  & \multicolumn{9}{c|}{\textbf{Accuracy (in $\%$) of CVD Prognosis Using Various Feature Selection Types}}\\ \cline{3-11} 
& & {$\beta$} & {$\alpha$-2}
& $\alpha$-4  & $\alpha$-6 & $\alpha$-8 & $\alpha$-10 & $\alpha$-12 & $\alpha$-14  & $\eta$ \\
\hline 
\hline

\multirow{5}[1]{*}{\textbf{Cleveland}}
 & LR & 95.1±3.2 & 87.0±3.0 & 95.3±1.9 & 92.9±2.3 & 97.7±0.4 & 95.3±2.0 & 97.6±0.4 & 94.1±3.2 & \textcolor{blue}{97.8±0.0}\\
\cline{2-11}
\centering & 
DT & 84.5±4.2 & 85.9±4.2 & 90.6±2.2 & 89.4±4.2 & 91.8±2.4 & 95.5±0.0 & 94.1±1.4 & 83.5±4.2 &\textcolor{blue}{95.9±2.2}\\
\cline{2-11}
\centering &
K-NN &81.7±3.2 & 89.4±1.7 &94.1±3.8 &91.6±3.2 &\textcolor{blue}{94.3±1.3}  &87.1±2.2 &70.6±3.8 &76.5±6.2 &67.5±6.1\\
\cline{2-11}
\centering &
NB &\textcolor{blue}{92.7±3.1} &58.8±1.9 &55.3±3.3 &82.4±2.3 &69.4±5.6 &75.3±5.3 & 90.6±0.9 &91.7±3.1 &75.3±2.1\\
\cline{2-11}
\centering &
SVM &98.1±02 &91.8±3.4 &96.4±2.5 &97.7±0.4 &98.8±0.9 &\textcolor{blue}{100.0} & \textcolor{blue}{100.0} & \textcolor{blue}{100.0} & \textcolor{blue}{100.0}\\
\hline
\hline

\multirow{5}[1]{*}{\textbf{Switzerland}} &
LR & 97.0±0.0 & 95.3±0.0 & 96.9±1.2 & 94.6±2.0 & 95.2±0.0 & 95.8±2.5 & 97.0±0.1 & 97.3±0.2 & \textcolor{blue}{97.7±0.2}\\
\cline{2-11}
\centering &
DT & \textcolor{blue}{96.8±1.0} & 91.9±0.9 &94.6±0.4 & 91.9±1.1 &96.3±0.9 &89.1±3.8 & 93.2±1.2 &91.9±2.4 &93.9±2.3\\
\cline{2-11}
\centering &
K-NN &95.4±1.0 & 93.9±1.1 & 92.4±1.2 & 90.0±0.0 & \textcolor{blue}{95.8±1.1} &86.5±5.0 & 94.9±0.9 & 94.7±2.7 &89.2±0.8\\
\cline{2-11}
\centering &
NB & 90.0±1.0 &\textcolor{blue}{91.9±0.8} & 89.2±2.0 & 91.8±0.6 & 67.6±4.5 &70.3±7.2 & 72.9±3.7 & 75.7±0.0 & 75.6±3.7\\
\cline{2-11}
\centering &
SVM &97.0±0.4 & 94.7±0.9 &91.9±3.2 & 95.0±1.9 & 97.0±0.4 &96.0±2.2 & 91.9±4.2 & 95.7±2.2 & \textcolor{blue}{98.9±0.1}\\
\hline\hline

\multirow{5}[1]{*}{\textbf{Hungarian}} &
LR & 96.8±1.1 & 88.8±2.7 & 89.9±0.0 & 93.3±2.3 & 97.2±1.1 & 95.3±1.1 & 97.8±0.9 & 96.6±2.1 & \textcolor{blue}{97.5±0.2}\\
\cline{2-11}
\centering &
DT & 93.2±3.0 & 86.5±4.1 & 93.3±1.4 & 86.5±3.2 & 85.4±3.3 &87.6±3.2 & 91.0±2.3 & 93.2±3.0 & \textcolor{blue}{94.1±2.7}\\
\cline{2-11}
\centering &
K-NN &91.0±4.2 & 94.4±0.0 &93.3±0.9 & 91.0±2.3 & 91.0±4.2 & 89.9±4.7 & 84.3±0.0 & 71.9±7.9 & \textcolor{blue}{94.9±3.3}\\
\cline{2-11}
\centering &
NB & \textcolor{blue}{92.1±3.8} & 85.4±3.2 & 84.3±2.2 & 87.6±1.1 & 92.1±1.0 & 91.0±1.8 & 82.2±4.3 & 92.1±3.9 & 80.9±7.3\\
\cline{2-11}
\centering &
SVM & 97.5±0.6 & 87.6±2.1 & 91.1±3.8 &93.3±3.1 &94.4±0.0 &96.6±2.7 &95.4±1.8 &98.0±0.4 & \textcolor{blue}{98.5±0.9}\\
\hline\hline

\multirow{5}[1]{*}{\textbf{Long Beach Va}} &
LR & 95.8±0.0 & 76.7±3.3 & 88.3±3.9 & 91.7±1.6 & 90.0±2.3 & 93.4±3.4 & 96.0±1.3 & 95.0±0.4 & \textcolor{blue}{96.3±0.4}\\
\cline{2-11}
\centering &
DT &89.0±1.9 & 80.0±2.3 & 81.7±1.2 & 83.4±3.2 & 86.9±3.2 & 91.87±2.5 & 88.4±2.5 & 87.0±1.2 & \textcolor{blue}{92.0±0.1}\\
\cline{2-11}
\centering &
K-NN & 80.0±2.9 & 70.0±2.8 & 83.3±2.7 & 85.0±1.7 & 81.4±5.3 & \textcolor{blue}{91.60±1.4} & 70.0±3.5 & 71.6±4.3 & 65.0±3.4\\
\cline{2-11}
\centering &
NB & 63.3±8.5 & 70.0±2.8 & 66.7±3.2 & 66.5±4.5 & \textcolor{blue}{80.0±0.0} & 63.3±4.4 & 58.9±6.5 & 63.3±8.5 & 60.0±5.8\\
\cline{2-11}
\centering &
SVM & 90.0±0.8 & 85.7±3.7 & 90.0±1.4 & 89.2±3.5 & 92.9±2.3 & 90.0±0.9 & 95.7±2.4 & 95.8±0.8 & \textcolor{blue}{97.0±1.4}\\
\hline\hline
\end{tabular}
\end{center}
\end{table*}

\subsubsection{Performance Analysis on The UCI Datasets}

Table~\ref{tab-result-ucis-datasets} compares the quantitative performances of five different classifiers on all four UCI datasets wrt different feature selection methodologies, including expert recommended features ($\beta$), ANOVA-based automatic top-n features ($\alpha$-$n$), and the proposed fusion-based feature selection ($\eta$). Similar to the analysis provided in Table~\ref{tab-result-BHDC}, this analysis also shows that the SVM model with the proposed feature fusion technique -- $\eta$ achieves the best CVD prognosis results with 100$\%$, 98.9±0.1$\%$, 98.5±0.9$\%$, and 97.0±1.4$\%$, respectively on the following datasets: Cleveland, Switzerland, Hungarian, and Long Beach Va dataset. In Table~\ref{tab-result-ucis-datasets}, each classifier's best result on each dataset is inked in \textcolor{blue}{blue}. 

\section{Conclusion} \label{sec:conclusion} 
 
It has been proved that the data-driven machine learning solutions are effective in predicting a disease possibility with an acceptable level of accuracy. Thus, such solutions can help to decrease the death rate, improve treatment planing, and reduce human errors in medical laboratory tests. However, it is found that there is a long way to go for the healthcare industry in this modern machine learning era, as there is always a necessity for a quick and accurate prognosis of cardiovascular diseases to safe lives. At this juncture, this work introduces a well-curated real data samples for cardiovascular disease research and develops an efficient model with excellent performance for cardiovascular disease prognosis. The experimental study on benchmark datasets shows that the proposed model is robust and achieves very competitive generalized performance compared to the existing approaches. 
The future work of this research is dedicated to: (i) the investigation of more predictor variables of cardiovascular diseases, novel feature selection techniques, and the application of advanced artificial neural network (ANN) models for cardiovascular disease prognosis, and (ii) the extension of the developed methodology for the prognosis of other diseases.

\section*{Acknowledgment}\label{sec:acknowledgment}
The authors would like to thank Dr. M. Shahabuddin, an Assistant Professor and the Head of the Department of Cardiology, Sylhet MAG Osmani Medical College, and Dr. Md. Anamur Rahman Anam for giving their valuable time to guide us in preparing the questionnaire and their support for data collection.

\bibliographystyle{IEEEtran}
\bibliography{reference.bib}

\end{document}